\documentclass[10pt,twocolumn,letterpaper]{article}

\usepackage[pagenumbers]{cvpr} 

\usepackage{graphicx}
\usepackage{amsmath}
\usepackage{amssymb}
\usepackage{booktabs}
\usepackage{times}
\usepackage{epsfig}
\usepackage{bm}
\usepackage{multirow}
\usepackage{cuted}
\usepackage{capt-of}
\usepackage{caption}
\usepackage{animate}
\usepackage{bbding}
\usepackage{footnote}
\usepackage{xcolor}
\usepackage{diagbox}
\usepackage{colortbl}
\usepackage{enumitem}
\usepackage{bmpsize}
\usepackage{bbm}

\makeatletter
\renewcommand\@makefnmark{\hbox{\@textsuperscript{\normalfont\color{red}\@thefnmark}}}
\makeatother

\usepackage[pagebackref,breaklinks,colorlinks]{hyperref}

\usepackage[capitalize]{cleveref}
\crefname{section}{Sec.}{Secs.}
\Crefname{section}{Section}{Sections}
\Crefname{table}{Table}{Tables}
\crefname{table}{Tab.}{Tabs.}

\newcommand{\bbm}{{\bf m}}

\graphicspath{{./images/}}

\definecolor{yellow}{rgb}{1, 1, 0.7}
\definecolor{orange}{rgb}{1, 0.85, 0.7}
\definecolor{pink}{rgb}{1, 0.7, 0.7}

\def\cvprPaperID{****} 
\def\confName{CVPR}
\def\confYear{2024}

\title{Disentangling Planning, Driving and Rendering for Photorealistic Avatar Agents}

\author{Duomin wang \quad Bin Dai \quad Yu Deng \quad Baoyuan Wang\\
	Xiaobing.AI \\
    \url{https://dorniwang.github.io/AgentAvatar_project/}
}

\begin{document}
\twocolumn[{
\renewcommand\twocolumn[1][]{#1}
\maketitle
\vspace{-30pt}
\begin{center}
    \captionsetup{type=figure} 
    \includegraphics[width=1\textwidth]{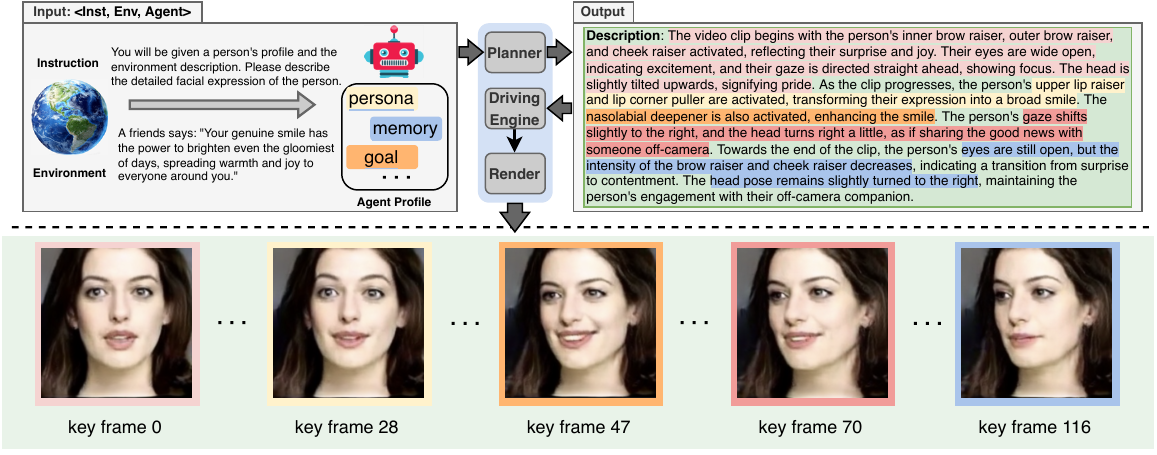}
    \caption{A showcase example. Our process begins with a high-level description of the environment and avatar, which is input into our LLM-based ``\textbf{planner}". This planner then produces a detailed description of facial motions. These descriptions are subsequently fed into the ``\textbf{driving Engine}" and, finally, the ``\textbf{render}" outputs photo-realistic video sequences that correspond to the initial inputs. }
    \label{fig:teaser}
\end{center}
}]

\begin{abstract}
\vspace{-3mm}
In this study, our goal is to create interactive avatar agents that can autonomously plan and animate nuanced facial movements realistically, from both visual and behavioral perspectives. Given high-level inputs about the environment and agent profile, our framework harnesses LLMs to produce a series of detailed text descriptions of the avatar agents' facial motions. These descriptions are then processed by our task-agnostic driving engine into motion token sequences, which are subsequently converted into continuous motion embeddings that are further consumed by our standalone neural-based renderer to generate the final photorealistic avatar animations. These streamlined processes allow our framework to adapt to a variety of non-verbal avatar interactions, both monadic and dyadic. Our extensive study, which includes experiments on both newly compiled and existing datasets featuring two types of agents – one capable of monadic interaction with the environment, and the other designed for dyadic conversation – validates the effectiveness and versatility of our approach. To our knowledge, we advanced a leap step by combining LLMs and neural rendering for generalized non-verbal prediction and photo-realistic rendering of avatar agents.

\end{abstract}

\section{Introduction}\label{sec:intro}
\textit{Autonomous Agents}~\cite{AgentRoadmap98,Agent_wang2023survey,xi2023rise} have experienced a resurgence in interest, both in industry and academia, thanks to the emergent capabilities of Large Language Models (LLMs). Recent open-source agents like Generative Agents~\cite{GA23}, AutoGPT~\cite{auogpt23}, Voyager~\cite{wang2023voyager}, and LLM-Planner~\cite{song2023llmplanner} exemplify LLMs' potential in sophisticated agent planning and decision-making. Using natural language as the interface is now considered paradigm shifting in the development of various autonomous agents~\cite{xi2023rise}. \textit{Avatars}, on the other hand, have been explored in various compelling scenarios for decades, including virtual presence, education, and social therapy, among others~\cite{VH97,InteractiveAvatar_CHI15,Social_Presence18,Social_interaction_survey22}. However, recent studies have predominantly focused on the visual ``representation" aspect of avatars~\cite{xiang2022gram,park2021nerfies,isik2023humanrf}, with their animation typically assumed to be driven by human-provided signals~\cite{Agent_Avatar_2020}. In the computer vision community, works such as video-based face reenactment~\cite{oorloff2023oneshot} and text or audio-based talking head generalization~\cite{Yu_2023_ICCV,Wang_2023_CVPR} adhere to this conventional paradigm, where input signals are expected to come from humans rather than an autonomous agent. We argue that these two research directions tend to be somewhat diverging, and there is a pressing need to bridge the gap between them. Specifically, how to leverage the latest breakthroughs in LLMs~\cite{xi2023rise} and neural representation~\cite{neural_rendering21} for developing advanced \textit{Autonomous Avatar Agents}?

By no means we are the first to marry LLMs with avatars for facial motion modeling in dyadic conversations,~\cite{Learn2Listen_LLM_2023_ICCV} fine-tunes a pre-trained LLM to autoregressively generate realistic 3D listener motion in response to the transcript of the speaker.~\cite{geng2023affective} also leverages LLMs for the mapping from the transcript of the speaker to the listener's facial attributes conditioning the goals of the listener, though both the driving and rendering are missing from this study. More early works on dyadic interactions also primarily focus on predicting the listener's behavior given the speaker's various inputs~\cite{PredictHeadPose_dyadic17,ahuja2019react,Chen2019ARF,Delbosc2023TowardsTG,Learning2Listen_2022_CVPR,ResponsiveHead_22,learning2smile_17}, text or audio. Conversely, in monadic interaction scenarios, the goal is to predict the motion behavior of the speaker given either audio~\cite{wav2Lip2020,zhou2021pose,Wang_2023_CVPR} or video~\cite{oorloff2023oneshot,zhang2021facial,ji2021audio} inputs. From a high-level perspective, all those previous works are scenario- and task-specific ad-hoc solutions rather than aiming for a generalized framework.

Before presenting our framework for developing avatar agents, we established several key design principles. First, the framework must offer the flexibility to craft avatar agents adept in both monadic and dyadic interactions. Second, it should possess the capability to execute plans based on high-level user instructions, which involve both reasoning and decomposition into small steps. Finally, for inherent generalization and scalability, it is crucial to separate high-level planning from the animation and rendering processes, ensuring that future enhancements in planning do not adversely impact other components. Adhering to these guiding principles, we have devised a novel framework encompassing three critical components:
\begin{enumerate}[topsep=0pt,itemsep=-1ex,partopsep=1ex,parsep=1ex]
    \item \textbf{LLM-Based Planner:} Given the high-level description of the environment, agent information such as ~\cite{NASS1995223}, it generates a sequence of text snippets, each providing a detailed textual description of the avatar's nuanced facial motions.
    \item \textbf{Task-agnostic Driving Model:} The driving engine transforms the detailed textual descriptions into a sequence of discrete facial motion tokens. 
    \item \textbf{Rendering Model:} These tokens are subsequently decoded into continuous facial motion embeddings (e.g., using VQ-Decoder~\cite{oord2018neural}). The rendering model then takes them as input and produces the final photorealistic animated avatar videos.
\end{enumerate}
We contend that our comprehensive framework is novel, with a specific emphasis on the first two components, which constitute our primary contributions. For simplicity, we utilize an existing rendering model from ``PD-FGC"~\cite{Wang_2023_CVPR}, although other similar rendering approaches are also applicable. It's worth noting that both the first and second components are LLM-based. Specifically, we first employ a pre-trained LLM as the planner, utilizing a carefully designed prompt structure as input, and fine-grained facial motion description as the desired output, which is a novel interface to drive avatar animations. Additionally, we fine-tune another LLM to convert the generated text into discrete facial motion tokens to build the driving engine. To verify the effectiveness of our framework, we carefully curated both the training and evaluation datasets and conducted extensive evaluations on top, both qualitatively and quantitatively. We plan to release the data and code for further research and development.

\section{Related Work}
\paragraph{Avatar Agents} Autonomous agents \cite{AgentRoadmap98, Human-level-Agent-18, Agent_wang2023survey} and avatars \cite{VH97, InteractiveAvatar_CHI15, Agent_Avatar_2020} have been subjects of extensive research for many years. The concept of an avatar agent \cite{mti21, Social_interaction_survey22} is crucial in various key applications like education, health therapy, social interactions, and gaming, especially when visual and behavioral aspects successfully cross the uncanny valley \cite{MACDORMAN2009695, Agent_Avatar_2020}. The level of automation, or ``agency", of these agents significantly impacts both productivity and user experience \cite{Social_interaction_survey22}. Before the advent of Large-scale Language Models (LLMs), avatar agents typically relied on human control for realistic experiences \cite{Agent_Avatar_2020}. Our work, diverging from previous approaches, revisits avatar agent design in light of recent advancements in LLMs \cite{xi2023rise, Agent_wang2023survey} and neural rendering \cite{neural_rendering21}. We introduce a novel framework that seamlessly integrates motion planning, driving, and photorealistic rendering in avatar agents, with text-based fine-grained motion descriptions and discrete motion tokens serving as the interface streamlining the three components.
\vspace{-2mm}

\begin{table}
\centering
\scalebox{0.8}{
    \begin{tabular}{c|ccc}
    \hline 
    Settings & Monadic & Dyadic or Triadic & Ours\\
    \hline 
    environment & \XSolidBrush & \Checkmark & \Checkmark \\
    persona & \XSolidBrush & \Checkmark & \Checkmark \\
    planning & \XSolidBrush & \XSolidBrush & \Checkmark \\
    animation & task-specific & task-specific & task-agnostic \\
    target & speaker & listener & both\\
    \hline 
    \end{tabular}
}
\caption{Comparison between our work with prior arts}
\vspace{-5mm}
\label{tab:related_work_compare}
\end{table}
\paragraph{Monadic, Dyadic and Triadic Interactions}
In the realm of neural-based video synthesis centered around speakers, we identify a category termed ``Monadic Interaction". This encompasses works like video-based single-image face reenactment \cite{oorloff2023oneshot,wayne2018reenactgan,zhang2021facial} and audio-based talking head generation \cite{zhou2021pose,Wang_2023_CVPR,wav2Lip2020}. Such studies focus on converting input signals to specific facial expressions, postures, or lip motions \cite{Yu_2023_ICCV,wav2Lip2020,Chu2018AFN,Jonell2019LearningNB,Ginosar_2019_CVPR,kucherenko2020gesticulator}, bypassing any high-level strategic planning. Typically, these models are developed in a task-specific manner. For instance, a 2D audio-to-face model is not transferable to a 3D talking face model due to their highly coupled driving and rendering processes, limiting scalability for broader applications. Meanwhile, works like \cite{ji2021audio,guo2021adnerf} developed user-specific models by somewhat overfitting the motion ``style" exhibited from videos, as opposed to explicitly modeling the agent's rich persona (expressed in text) of our work. On the other hand, non-verbal cues in dyadic \cite{Huang2017DyadGANGF,ahuja2019react,Learning2Listen_2022_CVPR, learning2smile_17} and triadic \cite{Joo_2019_CVPR} interactions are key in communicating information and emotions during human conversations. The typical objective is to generate listener motions that are compatible with the speaker's signals, often characterized by their non-deterministic nature \cite{Learning2Listen_2022_CVPR}. While early rule-based methods \cite{Virtual_Rapport,Virtual_Rapport2,bohus2010facilitating,Cassell1994AnimatedCR} struggled with generalizability, contemporary data-driven approaches \cite{learning2smile_17,Learning2Listen_2022_CVPR,Learn2Listen_LLM_2023_ICCV,ResponsiveHead_22,ResponsiveHead_22} have gained raising attention. A recent development \cite{Learn2Listen_LLM_2023_ICCV} involves fine-tuning a pre-trained LLM to translate speaker transcripts into discrete 3DMM motion tokens for the listener, utilizing the LLM's inherent knowledge and reasoning capabilities to address the challenge of non-deterministic mapping. Nevertheless, these existing methods remain task-specific in terms of animations and do not encompass high-level motion planning. Our approach, in contrast, employs an LLM-based planner for the heavy lifting of intelligence, while ensuring the animations are task-agnostic. This strategy allows our framework to be generalizable to both monadic and dyadic interactions and beyond. See Tab.~\ref{tab:related_work_compare} for a high-level comparison.
\vspace{-5mm}
\paragraph{LLM-based Planning}
Planning plays a vital role for agents to solve complex tasks. Chain-of-Though (CoT)~\cite{wei2022chain} and its variants, i.e., self-Consistent CoT~\cite{wang2022self}, Tree-of-Though~\cite{yao2023tree}, Graph-of-thought~\cite{besta2023graph}, design simple yet effective prompts such that the LLM can solve complex reasoning tasks step by step. ReAct~\cite{yao2022react} explicitly uses a planner to produce a sequence of thoughts and actions. Zero-Shot Cot~\cite{kojima2022large} demonstrates that LLMs are decent zero-shot reasoners given a proper prompt. Generative agents~\cite{GA23,wang2023humanoid} use a planning module to decide the daily schedule of an agent hierarchically. AutoGPT~\cite{auogpt23} uses planning as a controller to decide which model to use. Robotics Transformer~\cite{brohan2022rt,brohan2023rt} uses a planning module to generate high-level natural language instructions for the low-level robotics control. All those prior works use planner either to solve complex tasks~\cite{wei2022chain,wang2022self,yao2023tree,besta2023graph,yao2022react,kojima2022large} or to generate high-level schedules~\cite{GA23,wang2023humanoid}. In contrast, our work shows that LLMs can also be leveraged to predict nuanced human facial motions based on both environment and agent profiles.

\section{Approach}\label{sec:tech}
\begin{figure}
    \centering
    \includegraphics[width=1\linewidth]{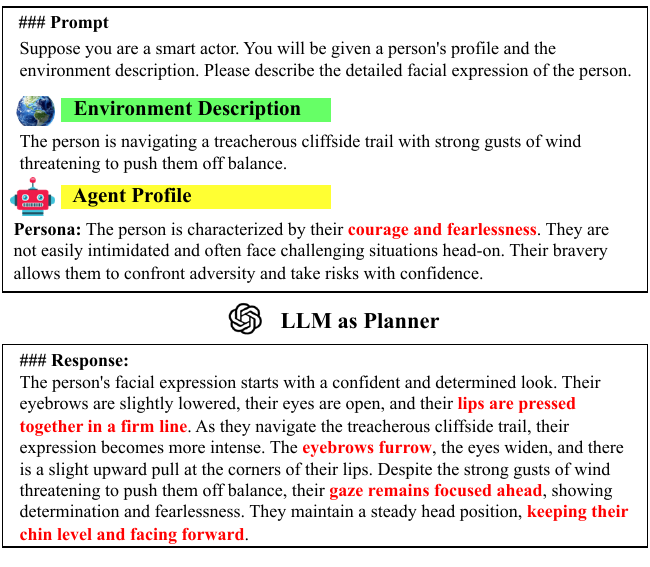}
    \caption{LLM-based Planner, which takes the information from both the environment and agent as input and generates the detailed text description of the facial motion for the avatar agent.}
    \label{fig:planner}
    \vspace{-4mm}
\end{figure}

As outlined in Sec.~\ref{sec:intro}, a fundamental principle of our design is to separate different modules and establish clear interfaces, ensuring that advancements in one module do not negatively affect others. With the recent breakthroughs in LLM-based agents~\cite{xi2023rise,GA23,auogpt23}, we find it compelling to steer the ``intelligence" aspect towards these LLM-based foundation models. This allows other components, like rendering, to concentrate on their specific functions. In pursuit of a significant advancement in this area, we introduce a framework comprising three distinct components: the LLM-based planner, the LLM-based task-agnostic driving engine, and the neural rendering engine. These are depicted in Figures~\ref{fig:planner},~\ref{fig:driving_module}, and~\ref{fig:render}, respectively. Figure.~\ref{fig:teaser} provides an overview of our framework, illustrating its process through a typical example.

\subsection{LLM-based Planner}\label{sec:planner}
Our LLM-based planner is responsible for perceiving the environment, interpreting the user's high-level instructions, and conducting essential reasoning to produce a detailed plan for facial motions that align semantically with the given inputs. Note that, this planning process is inherently non-deterministic, meaning there isn't a fixed input-output mapping. Nevertheless, the latest LLMs are deemed to possess certain emergent capabilities~\cite{wei2022emergent} that are likely to handle such complex non-deterministic planning. Even if not perfect, these capabilities are expected to improve alongside the scaling law~\cite{scaling_law} continuously. Thus, we rely heavily on the LLM-based planner for the majority of intelligent processing within our framework, aiming for better scalability. Intuitively speaking, LLMs already encode extensive knowledge, rich common sense, and even the World Model~\cite{gurnee2023language}, which can be leveraged for effective planning and reasoning.

In parallel, we recognize, akin to human perception, the necessity (and often sufficiency) of incorporating additional signals for more accurate and comprehensive planning. To enable the avatar agent to exhibit nuanced facial expressions and head poses, it is crucial to supply the planner with sufficient and discriminative information. Different from previous approaches that typically rely on a single, often insufficient condition—such as only the speech transcript of a speaker~\cite{Learn2Listen_LLM_2023_ICCV}—for predicting the listener's motion in dyadic conversations, our proposed structure holistically models a set of discriminative conditions, as elaborated below.

\vspace{-5mm}
\paragraph{Structured Conditions for Planner Input}
We define the Tuple \emph{S}: \texttt{<Inst,Env,Agent>} as the general input conditions for our LLM-based planner, where \texttt{Inst} denotes the user's instruction, \texttt{Env} denotes the environment the avatar agent is situated while \texttt{Agent} denotes anything related with the agent profile including persona. Figure.~\ref{fig:planner} shows one typical example of the input structure as well as the planned facial motion output by ChatGPT. It's worth noting that, thanks to the language interfaces, each text block of \emph{S} can be easily extended and generalized. For instance, in dyadic conversations, one can simply append the conversational history to \texttt{Agent} as part of its memory. 

\vspace{-5mm}
\paragraph{Guided Granularity for Planner Output} For In-context-learning~\cite{GPT-3} on top of LLMs, it's generally considered to be better if provided desired output format. In our scenarios, the coarse-grained description only includes the key information like \emph{``raised eyebrows, widened eyes"} while fine-grained information includes more details, like the example shown in Fig.~\ref{fig:planner}. To control the description granularity of the planner output, we can add some output examples in the prompt and leverage the instruction following and reasoning capability of the LLM. By this design, we can adapt the output of the planner to any kind of driving engine that takes a natural language as input without retraining a planner-specific driving engine. However, this part is intrinsically unnecessary if we choose to fine-tune the LLMs using pair-wised training data by specifying the text granularity directly in the desired output. We therefore omit this from the structured conditions for planner input, just for the sake of simplicity and clarity.  

\subsection{Task-Agnostic Driving Model}
\vspace{-1mm}
\subsubsection{Driving Model Disentanglement}
The goal of driving is to convert the text-based fine-grained motion description into discrete facial motion tokens. As illustrated in Fig.~\ref{fig:driving_module}, this driving model is also an LLM-based generative model that predicts the next motion token autoregressively conditioning on the text input. Technically, this can be regarded as one particular instruction-tuning task on top of pre-trained LLM~\cite{instructGPT22}. Different from those prior works~\cite{learning2smile_17,Learn2Listen_LLM_2023_ICCV,Learning2Listen_2022_CVPR,geng2023affective} which predominantly model the listener behavior given the speaker's speech transcript (or audio) in dyadic interaction, our driving model is agnostic to the avatar agent(either speaker or listener) and can be utilized in both dyadic and monadic interaction. This is because we let the planner(discussed in Sec.~\ref{sec:planner}) take care of the modeling of information related to the environment, dialogue history, and avatar personas while letting the driving model only handle a highly decoupled task, which only takes the textual description of the facial motions( without the identity) and transform into discrete motion tokens.
\vspace{-3mm}
\subsubsection{Driving Model Architecture}
\begin{figure}
	\centering
	\includegraphics[width=0.5\textwidth]{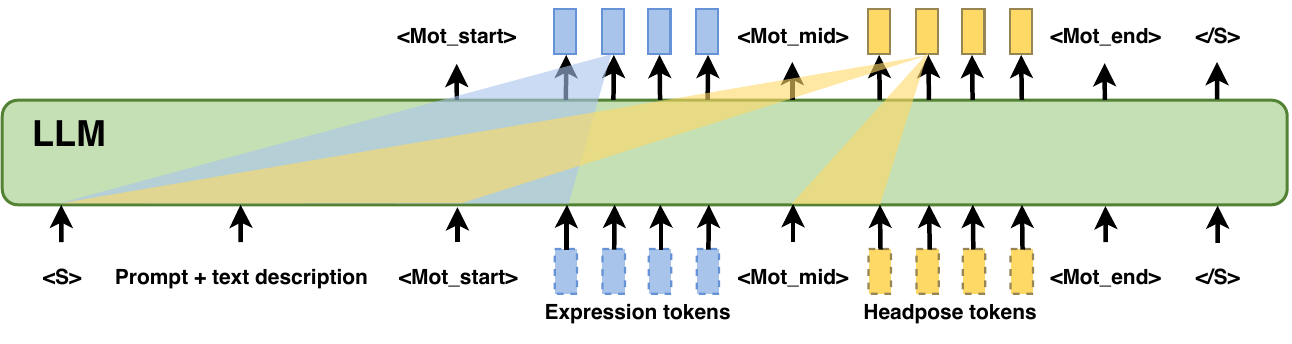}
	\caption{Our driving module, which converts a detailed textual description into discrete expression and head pose tokens.}
	\label{fig:driving_module}
\end{figure}
The facial motion in our context includes both non-rigid facial expressions and rigid head motions. We, therefore, train two VQ-VAEs~\cite{oord2018neural} to tokenize the embeddings extracted from the motion encoder of PD-FGC~\cite{Wang_2023_CVPR}, one for expression and eye-related embedding, and the other for head pose embedding. To enable LLM to predict those motion tokens, we have to extend the original language token vocabulary and resize the input embedding layer and the output head layer.  We additionally add up three special tokens, with \texttt{<Mot\_start>} and \texttt{<Mot\_end>} denoting the start and the end of either expression motion token sequences or headpose sequences, while \texttt{<Mot\_mid>} denoting their separation sign. We add extra attention masks to avoid dependence on expression tokens when predicting the tokens for head pose on top of the normal causal transformer. Figure.~\ref{fig:driving_module} illustrates the masking mechanism and the special tokens.

\subsubsection{Building Fine-grained Text-to-motion Dataset}\label{sec:fine-grained-data}

To train the driving model autoregressively, we have to build a pair-wise dataset with such format \texttt{<fine-grained text description of facial motion, discrete motion tokens>}. For this purpose, we developed an automatic data curation pipeline through unsupervised videos, by unitizing face-understanding models such as expression extraction and pose estimations, and then converting them into textural descriptions by leveraging the emergent ability of LLMs, such as ChatGPT.  Specifically, we built upon Celebv-text~\cite{celebv_text_cvpr23} from which we randomly choose $4,300$ videos, $4,000$ for training, and the rest for testing. We leverage the following methods to obtain fine-grained facial motions that include facial muscles, eye blink, eye gaze, and head pose. We name our dataset as \textbf{Fi}ne-grained Ce\textbf{l}ebv \textbf{M}otion Description \textbf{Data}set(short for \textbf{``FilmData"}).

\begin{figure}
    \centering
    \includegraphics[width=0.99\linewidth]{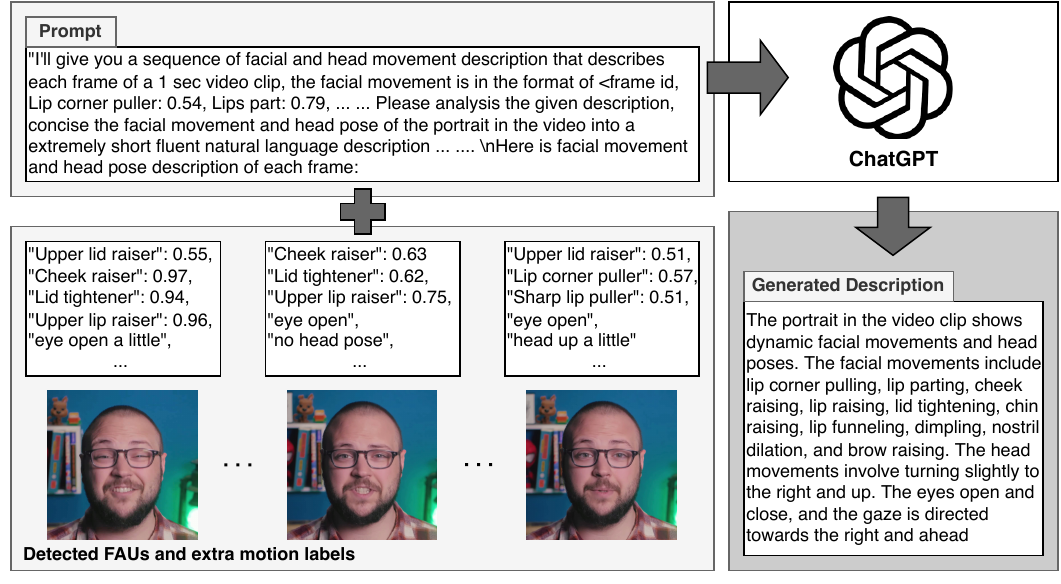}
    \caption{An example illustrating the process of converting FAUs and extra motion labels into textual descriptions via ChatGPT.}
    \label{fig:fine-description-generate}
    \vspace{-3mm}
\end{figure}

\vspace{-3mm}
\paragraph{Facial Action Unit Detection.} We first take advantage of an off-the-shelf facial action unit detector~\cite{me_graphau} to get 41 FAUs results for each frame of the videos. The detected FAUs include most of the facial muscle movements, and each FAU has a text-based descriptive label, such as \textit{inner brow raise} and \textit{lip corner pull}. We set FAUs as activated if their probability is greater than 0.5 for each frame.
\vspace{-3mm}
\paragraph{Motion Embedding Cluster.} Although FAUs have detailed facial movement descriptions, datasets~\cite{mavadati2013disfa, zhang2014bp4d} that are used to train FAU detectors lack of eye blink, eye gaze, and head pose labels. Thus we design a new method to label these attributes through GMM-cluster based on a pre-trained motion encoder from PD-FGC~\cite{Wang_2023_CVPR}. Concretely, we extracted disentangled motion features of each frame using the motion encoder mentioned above. Then we conduct a GMM-cluster on eye blink features, eye gaze features and head pose features, respectively. We label the clustering result manually and get 9 classes for gaze, 9 classes for the head pose, and 5 classes for an eye blink.
\vspace{-3mm}
\paragraph{From Attributes to Text Description.} Once we obtain 41 FAUs and 23 extra motion labels for each frame, we feed these attributes of 25 consecutive frames into ChatGPT together with carefully designed prompts to generate descriptions for a 1-second video clip, see Fig.~\ref{fig:fine-description-generate}. To generate descriptions for longer video clips, we can leverage the results of a 1-second video to form a longer description, such as 5 seconds, and then feed them into ChatGPT together with another prompt for longer video description generation. We create descriptions for long video clips in this manner to circumvent the maximum token limitation that would be exceeded if we were to directly input the results of multiple consecutive frames into ChatGPT.

Refer to the supplemental materials for more details regarding the data sets as well as FAUs and motion labels.

\subsection{Rendering Model}
\begin{figure}
    \centering
    \includegraphics[width=0.99\linewidth]{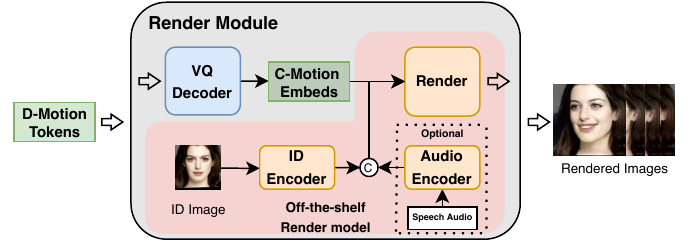}
    \caption{Illustration of the major components within our render. Audio input enables when agent act as ``speaker"}
    \label{fig:render}
     \vspace{-3mm}
\end{figure}

Given the discrete motion tokens output by the driving module, our render first transforms them into continuous motion embeddings through VQ-Decoder ~\cite{oord2018neural}. Without losing generality, we employ an off-the-shelf rendering model for one-shot talking head, namely PD-FGC~\cite{Wang_2023_CVPR}, further to convert the motion embeddings into final photorealistic videos. For the sake of clarity, and thanks to the disentangle ability of PD-FGC, we make speech audio as optional input, while primarily relying on the motion embeddings to control the final rendering for any given identity image. Fig.~\ref{fig:render} illustrates the major components within our render. For more details about PD-FGC's work, please refer to the original paper~\cite{Wang_2023_CVPR}.

\section{Experiments}\label{sec:experiment}
We first provide a detailed evaluation of the planner in Sec.~\ref{sec:planner_evaluation}. To thoroughly evaluate the driving engine, we proposed a set of baselines and evaluated it from three different spaces including the emotion embedding space, 3DMM space as well and the text space defined by the generated motion descriptions. We also conducted end-to-end evaluations at the video level in text description space. Lastly, we provided a user study comparing with a SOTA system ~\cite{Learn2Listen_LLM_2023_ICCV}.

\subsection{Planner Evaluation}\label{sec:planner_evaluation}
We conduct experiments on two typical cases to evaluate our LLM-based planner, one for the dyadic conversation between two agents and the other for a single-agent monadic interaction with the environment, though our planner can be applied for broader scenarios as discussed in Sec.~\ref{sec:planner}.

\vspace{-3mm}
\paragraph{Case 1: Dyadic Conversation.} We evaluate the performance of non-verbal behavior prediction for both \emph{listener} and \emph{speaker} in dyadic conversations. When modeling the listener, the environment information is the transcript of the speaker and vice versa, which essentially degenerates to the settings of~\cite{geng2023affective,Learn2Listen_LLM_2023_ICCV} when no other information is considered. Furthermore, the conversation history can be regarded as the short-term memory of the listener, providing more clues for facial motion decisions. As for the speaker, both the transcript and the conversation history can be regarded as the agent's information.\footnote{Other important environmental information is also possible to be considered, such as the relationship between the speaker and the listener, and the location where the conversation happens, to name a few.}
We use the test set of \texttt{DailyDialogue}~\cite{li2017dailydialog} to evaluate the planner which contains 1000 multi-turn conversations covering various topics in daily life between two persons. Each turn is labeled with the emotion of the speaker. We only use the sentences that have at least $3$ previous turns such that each test sentence has enough conversation history. In addition, to avoid trivial sentences that lead to neutral facial expressions, we only select the sentences whose corresponding emotion is not \emph{neutral}. A total of $786$ data samples are thereby obtained.
\vspace{-3mm}
\paragraph{Case 2: Monadic Interaction with Environment.} In this case, the planner is still expected to predict the non-verbal behavior of the agent when only an environment description is provided, with no other interacting agents. For example, when situated in an environment described like ``\emph{The person is alone in a haunted house, hearing strange noises and seeing objects move on their own}.", though no conversation happens, one can imagine how the agent with a different persona would behave non-verbally. For this evaluation, we synthesize a dataset with both diverse environment descriptions and various personas, resulting in a $1200$ combination of (environment, persona) pairs. We name the synthesized dataset as ``\texttt{EnvPersona}". The detailed synthesizing process can be found in the supplementary file.

\begin{figure}
    \centering
    \includegraphics[width=1.0\linewidth]{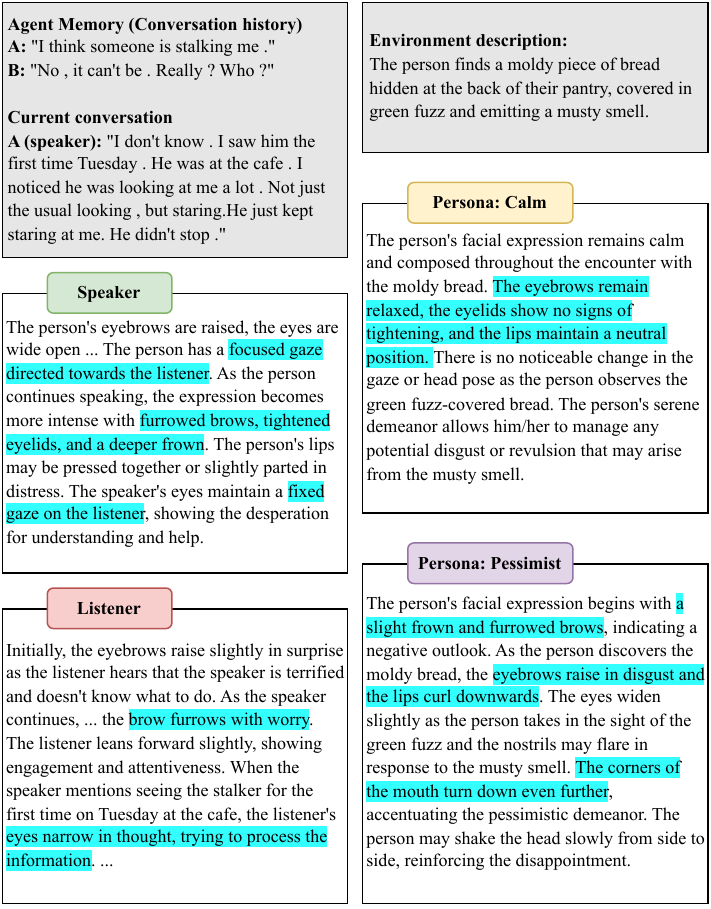}
    \vspace{-3mm}
    \caption{The qualitative results of the planner from (left) \texttt{DailyDialogue} and (right) \texttt{EnvPersona} datasets.}
    \label{fig:planner_gallery}
    \vspace{-5mm}
\end{figure}

\vspace{-3mm}
\paragraph{Qualitative Results.} We show two randomly selected results for qualitative evaluations in Fig.~\ref{fig:planner_gallery}. The left example is from the \texttt{DailyDialogue}. When person A tells person B she was stalked, the planner produces reasonable descriptions of both persons. The speaker has \emph{``furrowed brows, tightened eyelids, and a deeper frown"}, which is plausible for a worried person. On the other hand, the listener, though also worried, has \emph{``eyes narrow in thought"}, meaning that he is \emph{``trying to process the information"} of the stalking. The right is an example from the \texttt{EnvPersona}. We can see that persons with different personas express differently even if they are situated in the same environment.

\begin{figure*}[t]
	\centering
	\includegraphics[width=0.95\textwidth]{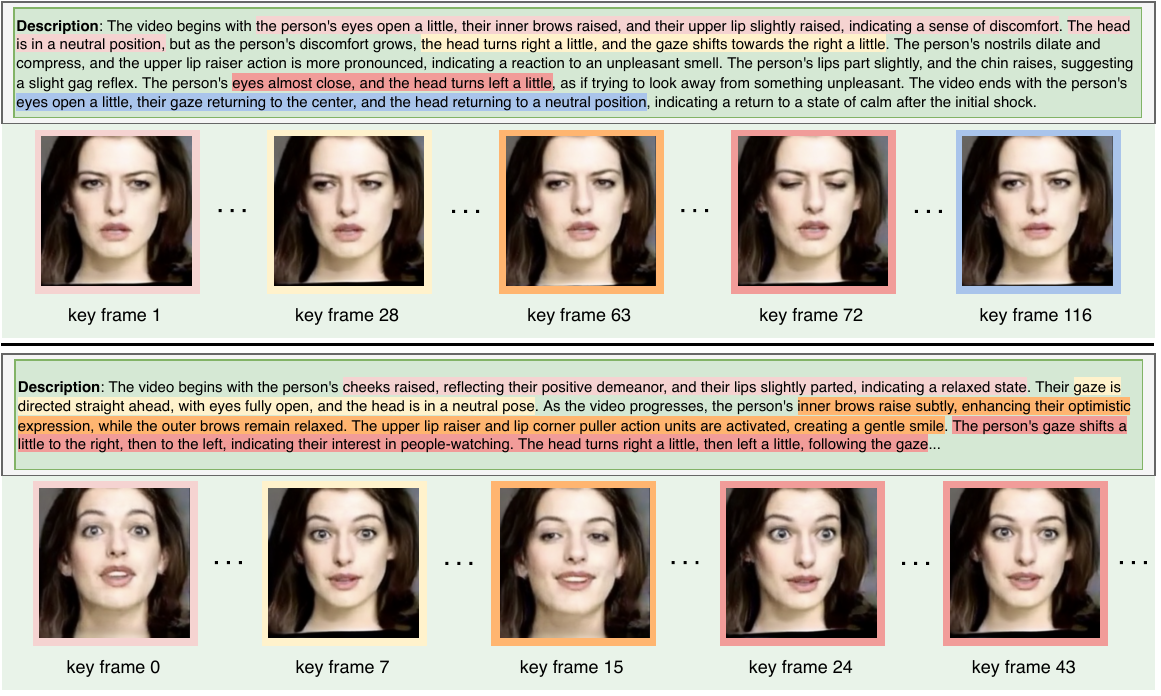}
        \vspace*{-2mm}
	\caption{The qualitative result of driving engine. We show two different cases here to demonstrate the text-driven capability of our model.}
	\label{fig:driving_gallery}
\end{figure*}

\begin{table*}[htbp]\small 
    \centering
    \begin{tabular}{c|ccc|ccc|ccc}
    \toprule
    & \multicolumn{3}{|c|}{DailyDialogue-Speaker} & \multicolumn{3}{|c|}{DailyDialogue-Listener} & \multicolumn{3}{|c}{EnvPersona-Agent} \\
    \hline
    Hit@ & Hit@1 & Hit@5 & Hit@10 & Hit@1 & Hit@5 & Hit@10 & Hit@1 & Hit@5 & Hit@10 \\
    \hline 
    Random & 5.0 & 25.0 & 50.0 & 5.0 & 25.0 & 50.0 & 5.0 & 25.0 & 50.0 \\
    No memory/persona & 41.0 & 77.5 & 94.5 & 27.0 & \textbf{66.5} & 87.5 & 37.2 & 73.3 & 86.9 \\
    Ours & \textbf{45.5} & \textbf{87.0} & \textbf{96.0} & \textbf{29.5} & \textbf{66.5} & \textbf{90.0} & \textbf{45.0} & \textbf{93.2} & \textbf{100.0} \\
    \bottomrule  
    \end{tabular}
    \vspace{-2mm}
    \caption{Quantitative evaluation of the planner. Best performance is achieved when both environment and agent information are included. Removing the memory/persona of the agent degrades the performance.}
    \label{tab:planner_hit_at_k}
   \vspace{-2mm}
\end{table*}

\vspace{-3mm}
\paragraph{Quantitative Results.} We employ the \emph{Hit@$k$} metric for quantitatively assessing the planner's performance through relative comparison. To be specific, we first run our planner for all test data to get their fine-grained motion descriptions as planning results. For each test example, we then randomly sample additional $\bbm$ descriptions from the remaining predicted descriptions in the entire dataset. Next, we leverage another LLM (i.e., GPT-4\footnote{GPT-4 is used only for evaluation, by default we use GPT-3.5}) as the critic to rank all $\bbm+1$ descriptions based on the semantic likelihood, from the most probable to the least. 
If the true predicted description's rank is within the top $k$, we consider the sample as a successful hit at $k$. The reported metric is the ratio of samples that achieve a correct hit at $k$. Throughout our evaluations, we consistently set $\bbm$ to 19. Due to space limits, refer to supplemental material for more details regarding how this metric is computed. Tab.~\ref{tab:planner_hit_at_k} shows the numeric results on both datasets. We compare our method with the \emph{Random} baseline to conceptually verify if the SOTA LLM can plan facial expressions.  We also conducted ablations. For \texttt{DailyDialogue}, we remove the conversation history; and for \texttt{EnvPersona}, we remove the persona of the agent. Surprisingly, our results on \emph{Hit@$5$} and \emph{Hit@$10$} are quite encouraging, indicating the strong planning capability exhibited in those LLMs. Meanwhile, when the persona or memory is removed, the performance is degraded consistently in all settings, demonstrating that both information are essential during planning. Nevertheless, the \emph{Hit@$1$} results still seem low. There are two possible reasons behind this. First, the result itself is by nature non-deterministic, a fixed sequence of facial motions may apply to multiple environmental inputs, which hurts \emph{Hit@$1$} metric. Second, our current implementation only leverages the in-context learning of LLMs, future dedicated fine-tuning would likely improve the performance.

\begin{table*}[th!]\small
\centering
\begin{tabular}{l|cc|cccc|cccc} 
\toprule
\multicolumn{1}{c}{\multirow{2}{*}{Method}} & \multicolumn{2}{c}{Text} & \multicolumn{4}{c}{3DMM Parameters(after render)}  & \multicolumn{4}{c}{Motion Embeddings(before render)}        \\
\cline{2-11}
\multicolumn{1}{c}{} &
$\text{MSP} \uparrow (\%)$ &  
$\text{MSE} \uparrow(\%)$ & 
$\text{Var}  \rightarrow$ &
$\text{FID}\downarrow$ & 
$\text{FID}_{\Delta}\downarrow$ &  
$\text{SND}\downarrow$ &
$\text{Var}  \rightarrow$ &
$\text{FID}\downarrow$ & 
$\text{FID}_{\Delta}\downarrow$ &  
$\text{SND}\downarrow$ 
\\
\midrule
  GT &   100&  100 & 2.24 & - & - & - & 14.66 & - & - & -\\
  \midrule
  $\text{Meaningless}$ &    0.0  & 0.0  & -  & - & - & - & - & - & - & -  \\
  $\text{Average}$ &  49.5 &  56.7 &  -  &  -  &  -  &  - & - & - & - & -\\
  $\text{Random}_{Gen}$ & 64.81  & 66.16  &  \textbf{1.96}  &  8.77  &  0.12  &  8.89 & \textbf{12.34} & 67.19 & 0.74 & 67.93\\
  $\text{Random}_{GT}$ & 59.62  & 60.23  &  -  &  17.07  &  0.13  &  17.20 & - & 66.17 & 0.61 & 66.78\\
\midrule
$\text{Ours}$ &  \textbf{73.32}  &  \textbf{71.63}  &  1.65 & \textbf{7.08}  &  \textbf{0.10}  & \textbf{7.18}  & 10.96 & \textbf{37.20} & \textbf{0.49} & \textbf{37.69} \\
\qquad  w/o $LoRA$& 67.05  &  68.81 & 1.71  &  8.10 & 0.11 & 8.21 & 11.56 & 56.34 & 0.64 & 56.98 \\
  \bottomrule 
\end{tabular}
\vspace*{-2mm}
\caption{The quantitative results of text-motion matching, motion variance, and naturalness for LLM \textbf{generate} motion sequence. "$\rightarrow$" indicates a closer score to GT is better.}
\vspace*{-4mm}
\label{table:text_motion_matching}
\end{table*}


\subsection{Driving Model Evaluation}

\paragraph{Dataset and Implementation Details} We train our driving model on the dataset proposed in Sec.~\ref{sec:fine-grained-data}, which includes $4,000$ videos that further split into $8,678$ 5-second clips. The testing set includes $300$ 5-second clips randomly sampled from the original $300$ testing videos. We chose Llama~\cite{touvron2023llama} 7B as the LLM backbone and fine-tuned a LoRA~\cite{hu2021lora} together with the input embedding layer and output head layer. The optimizer is Adam~\cite{kingma2014adam} with learning rate 4e-4, weight decay 1e-3, betas 0.9 $\&$ 0.95. The batch size is set as 4 for each GPU, and the gradient accumulation step is 4. The model was trained on 4 NVIDIA A100 GPUs for about 20 hours for a total of 60 epochs.
\paragraph{Evaluation Space}
We evaluate the driving engine from three spaces, including the motion embedding space before the rendering, the 3DMM parameter space (extracted from videos) after rendering, and the text space (i.e., generated from video as sec.\ref{sec:fine-grained-data}).

\vspace{-3mm}
\paragraph{Baselines.} 
We compare our method with the following baselines:
\textbf{1) GT}, represents the ground truth, and \textbf{Meaningless}, represents sentences that have nothing with facial movement and head pose. For example, ``\textit{Essays are commonly used as literary criticism, political manifestos, learned arguments...}".  We randomly select those sentences from Wikipedia to form the Meaningless set. These two baselines serve as a reference to verify the effectiveness of GPT-4(which is the critic) and indicate the variance of ground truth motion.
\textbf{2) Average}, we average the motion embedding sequence of the training set and compute the averaged result with testing set ground truth in text space.
\textbf{3) $\text{Random}_{Gen}$}, we fit a multivariate Gaussian distribution from the motion embedding space and generate results by sampling from it, and \textbf{$\text{Random}_{GT}$}, we compare the generated result with a randomly chosen ground-truth from the test set. All three spaces are evaluated for these two baselines, and we ran these two baselines 10 times and used the average score as the final score.

\begin{table}[htbp]
    \small
    \centering
    \begin{tabular}{c|c|c|c}
    \toprule 
    & DailyS & DailyL & EnvPersona \\
    \hline
    Hit@ & Hit@3  & Hit@3  & Hit@3 \\
    \hline 
    Random & 30.0 & 30.0 & 30.0  \\
    Ours & \textbf{37.3}  & \textbf{36.8} & \textbf{35.6}  \\
    \bottomrule  
    \end{tabular}
    \vspace*{-2mm}
    \caption{Quantitative evaluation of the entire system.}
    \label{tab:system_hit_at_k}
   \vspace*{-3mm}
\end{table}

\begin{table}[htbp]
    \small
    \centering
    \begin{tabular}{c|c|c|c}
    \toprule 
     Method & $\text{Naturalness}\uparrow$ & $\text{Matchness}\uparrow$ & $\text{Var}\uparrow$ \\
    \hline 
    LM-Listener~\cite{Learn2Listen_LLM_2023_ICCV} & 30$\%$ & 22$\%$ & 5$\%$  \\
    Ours & \textbf{45$\%$}  & \textbf{58$\%$} & \textbf{72$\%$}  \\
    \bottomrule 
    \end{tabular}
    \vspace*{-2mm}
    \caption{User study between ours with others.}
    \label{tab:user_study}
   \vspace*{-4mm}
\end{table}

\vspace{-3mm}
\paragraph{Quantitative Results.} We evaluate our method with the following metrics:
\textbf{(1) MSP} and \textbf{MSE} measure the matching degree of head pose and facial expression between the input motion description and generated motion sequence. Directly computing matching degree is non-trivial, we conduct this by first producing a rendered video from the predicted motion sequence, then generating detailed text description from the rendered video through the method we proposed in Sec.~\ref{sec:fine-grained-data}, and finally using ChatGPT to score the matching degree in time order of input motion description and extracted motion description with a carefully designed prompt, the scoring procedure is performed independently for head pose and facial expression.
\textbf{(2) }The naturalness of generated motion sequence after rendering: \textbf{Var}, \textbf{FID}, \textbf{$\text{FID}_{\Delta}$} and \textbf{SND}, are borrowed from ~\cite{Yu_2023_ICCV}, and computed on both 3DMM coefficients extracted from ~\cite{deng2019accurate} and generated motion embeddings. The results are shown in Tab.~\ref{table:text_motion_matching}. The \textbf{MSP} and \textbf{MSE} score of GT is $100\%$, which means an exact match of head pose and facial expression, \textbf{Meaningless} scores 0 on both \textbf{MSP} and \textbf{MSE}, which mean completely mismatch. Our method outperforms $\text{Random}_{GT}$ by nearly 10 percent in text space. Our method also performs best on naturalness metrics and is comparable with $\text{Random}_{Gen}$ on \textbf{Var}, even though $\text{Random}_{Gen}$ is fitted to the distribution of the training set.
\vspace{-3mm}
\paragraph{Qualitative Results.} We conduct qualitative evaluation through Fig.\ref{fig:teaser}, \ref{fig:driving_gallery}, and those in supplemental materials, all indicating the effectiveness of our method.

\subsection{End-to-end System Evaluation}
We perform the whole-system evaluation at the video level in the text description space. We first run the system from end to end to obtain the final rendered videos for each input, we then extract motion descriptions from each rendered video following the procedure introduced in Sec.~\ref{sec:fine-grained-data}. Finally, we use the same metric proposed in Sec.~\ref{sec:planner_evaluation}, namely, \emph{Hit@$k$}, to measure semantic matching at the video-level descriptions. The results are shown in Tab.~\ref{tab:system_hit_at_k}. Note that, aside from the non-deterministic nature which might cause certain ambiguity for ChatGPT to distinguish, another reason causing the low \emph{Hit@$3$} is the accumulated error along our end-to-end evaluation pipeline. Coming up with a better metric is still an open challenge.

\subsection{Compared with Other Methods}
We compared our system with ~\cite{Learn2Listen_LLM_2023_ICCV} under the \textit{Listener} situation through a user study based on three aspects: \textbf{1) Naturalness}  and \textbf{2) Variance} of generated facial motion, and the \textbf{3) Matchness} between the generated motion and input context. We use Vico Dataset~\cite{ResponsiveHead_22} as the test set for this comparison. Tab.~\ref{tab:user_study} shows that our method got significantly better performance in all three metrics. More details are in supplementary materials.

\vspace{-2mm}
\section{Conclusion and Future Works}
We have developed a streamlined yet disentangled pipeline designed to predict the non-verbal behaviors of a photo-realistic avatar agent, applicable in both monadic and dyadic interactions. While the outcomes are promising, achieving a fully autonomous avatar agent capable of processing text, video, and audio inputs, and predicting not just non-verbal facial behaviors but also comprehensive body movements and verbal communication, remains a significant challenge. However, the generality and adaptability of our disentangled principle and overall framework lay a strong foundation for future advancements in this area.

{\small
\bibliographystyle{ieee_fullname}
\bibliography{egbib}
}

\clearpage
\appendix

\begin{strip}
\centering
\Large{\textbf{Supplementary Material}}
\end{strip}

\renewcommand{\thesection}{\Alph{section}}
\renewcommand{\thefigure}{\Roman{figure}}
\renewcommand{\thetable}{\Roman{table}}
\renewcommand{\theequation}{\Roman{equation}}
\setcounter{figure}{0}
\setcounter{equation}{0}

\section{Overview}
In this supplementary material, we give more details about the prompt design, data generation, and module design in Sec~\ref{sec:implentation_details}. We also describe details about the evaluation protocol in Sec.~\ref{sec:evaluate}, and show more results from different modules in our system in Sec.~\ref{sec:results}. Meanwhile, in Sec.~\ref{sec:limitation}, we further provide discussions regarding the limitations of the current system and offer potential future works for improvement, encouraging more research effort along this line.

\section{Advantages of Disentangled Design}\label{sec:disentangle}
We argue that the disentanglement design of our system offers the following important advantages.
\paragraph{Generalized for Multiple Tasks}
The most significant advantage is that our system can competent many different tasks without re-training. For example, our system can act as a ``speaker" as well as a ``listener" when in a conversation situation, or just interact with the environment with a series of self-controlled behaviors. This enables our system to generalize to different tasks and variant scenes, extremely reducing the memory cost and time consumption as well as the complexity of integration when running as a part of a self-controlled avatar system.
\paragraph{Develop/Improve Independently}
The overall disentangled design enables the major components to develop on their own leveraging their respective breakthroughs. For example, improving the planner using a more advanced LLM(with better planning capabilities) would make the final results better even without changing the driving and rendering engines. Likewise, upgrading the rendering to make it more 3D-aware and with high-resolution rendering quality would still work for both the planner and the driving engine. For example, it's highly expected to get better final rendering results by combining the high-resolution 3D consistent rendering,i.e., GRAMHD \cite{xiang2022gram}, even with the current motion token spaces.

\paragraph{Low Implementation Cost}
Another benefit of disentanglement is that it can extremely reduce the implementation cost of the entire system. The rendering module we use is from an off-the-shelf model, so we do not need to implement it from scratch. The driving module and planner can be implemented independently, so we do not need to collect pair data, which is motion sequence and multiple inputs(including personality, environment, memory and \etc.) of the planner.

\section{Evaluation Protocal}\label{sec:evaluate}
We'd like to emphasize that evaluating each component in our system is a non-trivial task. Here, we describe how we conduct evaluation for each component and analyze the limitations of the evaluation pipeline for the entire system.
\subsection{Evaluation Pipeline for Planner}

\begin{figure}
    \centering
    \includegraphics[width=0.9\linewidth]{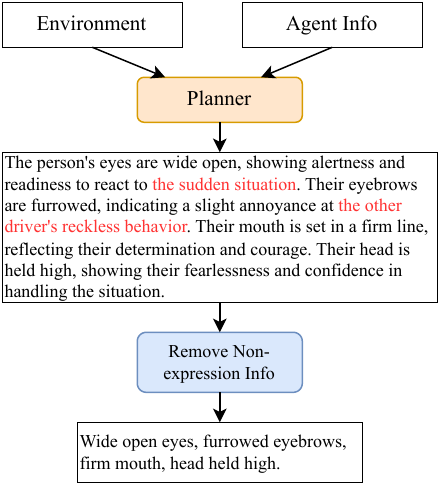}
    \caption{Non-expression part removal in fine-grained description.}
    \label{fig:clean_expression}
\end{figure}

As outlined in the main paper, we employ the \emph{Hit@k} metric to assess the planner's performance quantitatively. However, a noteworthy caveat exists, where the final detailed descriptions from the planner may inadvertently incorporate information from either the environment or the agent. This reliance on extraneous details can compromise the evaluation process, leading to ``information leakage". An illustration is depicted in Fig.~\ref{fig:clean_expression}, where the highlighted details reveal information from the environment by referencing \emph{the other driver} in the description.

To mitigate the information leakage issue, we employ a filtering process using ChatGPT to exclude contents related to the descriptive information from the environment or the agent. Consequently, only contents pertaining to facial expressions are retained after this filtration. It is acknowledged that this filtering procedure may potentially exclude some valuable fine-grained expression information, thereby affecting the evaluation performance. Nevertheless, the achieved performance remains promising, as demonstrated in Tab.~\ref{tab:planner_hit_at_k} of the main paper.

\subsection{Evaluation Pipeline for Driving}\label{sec:driving_evaluate}
We conduct evaluations for the driving module in the following three feature spaces: motion embedding space, 3DMM space, and text space. 

For the motion embedding space, we first convert the generated motion tokens into continuous motion embeddings, and then compute \textbf{Var}, \textbf{Fid}, \textbf{$\text{FID}_{\Delta}$} and \textbf{SND} with ground truth embeddings.

As for 3DMM space, the generated motion tokens are fed into the render module to generate corresponding videos first, then we extract 3DMM parameters from generated videos and ground truth videos, and then compute \textbf{Var}, \textbf{Fid}, \textbf{$\text{FID}_{\Delta}$} and \textbf{SND} between them.

In order to evaluate text-motion matching degree, we need to get fine-grained motion description text and evaluate in text space. The overall pipeline to obtain descriptions for generated motion tokens consists of four steps: 
\begin{itemize}
    \item Render the generated motion tokens into videos.
    \item Extract FAUs and extra labels from the rendered videos.
    \item Translate the extracted attributes into 1-second descriptions.
    \item Combine 1-second descriptions to form a longer, such as 5-second, motion description.
\end{itemize}  
After we get the description text of the generated token sequence, we utilize ChatGPT to score the motion matching degree between the generated description and ground truth description in time order and take the average score from all results. This evaluation is performed independently for head pose and expression.
\subsection{Evaluation Pipeline for the End-to-end System}\label{sec:system_evaluate}
For the entire system evaluation, we feed the environment into the system and obtain generated motion tokens. We evaluate the entire system using the same \emph{Hit@k} metric as the evaluation of the planner. But in order to compute \emph{Hit@k}, we need to translate the generated motion tokens into text descriptions. The first four steps are the same as text space evaluation mentioned in~\ref{sec:driving_evaluate}. Besides, we extract a concise motion phrase from the generated fine-grained description to prevent information leakage and then compute \emph{Hit@k} using the extracted motion phrase.

\begin{table*}
    \centering
    \begin{tabular}{c|ccc|ccc|ccc}
    \hline 
     & \multicolumn{3}{|c|}{DailyDialogue-Speaker} & \multicolumn{3}{|c|}{DailyDialogue-Listener} & \multicolumn{3}{|c}{EnvPersona-Agent} \\
     \hline 
     Hit@ & Hit@1 & Hit@2 & Hit@3 & Hit@1 & Hit@2 & Hit@3 & Hit@1 & Hit@2 & Hit@3 \\
     \hline 
     Random  & 10.0 & 20.0 & 30.0 & 10.0 & 20.0 & 30.0 & 10.0 & 20.0 & 30.0  \\
     Ours & \textbf{13.2} & \textbf{27.7} & \textbf{37.3} & \textbf{11.7} & \textbf{27.9} & \textbf{36.8} & \textbf{12.3} & \textbf{23.2} & \textbf{35.6} \\
     \hline 
    \end{tabular}
    \caption{Additional quantitative evaluation of the entire system.}
    \label{tab:system_results}
\end{table*}

\section{More Implementation Details}\label{sec:implentation_details}
\subsection{Prompt Design of Planner}
\begin{figure*}
    \centering
    \includegraphics[width=0.99\textwidth]{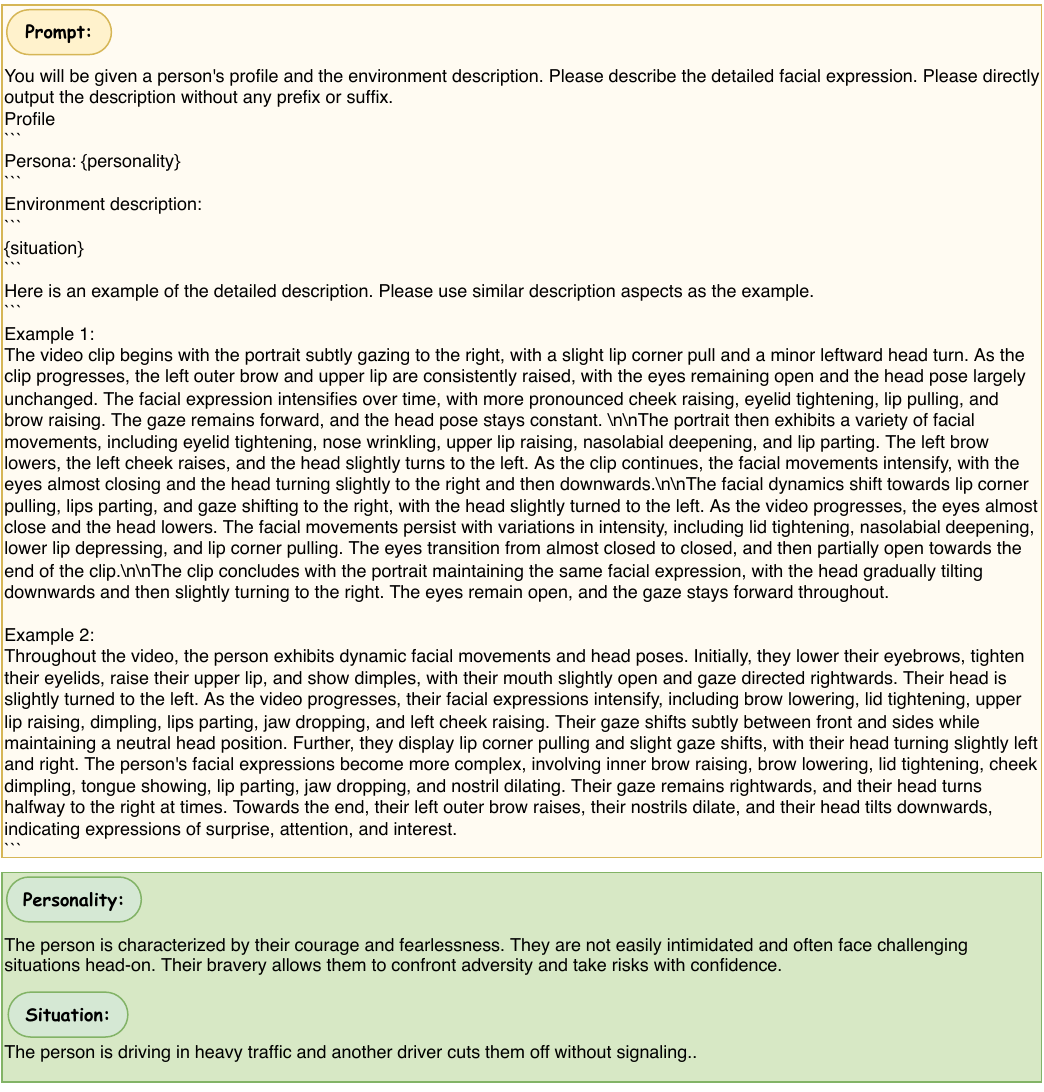}
    \caption{Planner prompt for \texttt{EnvPersona}.}
    \label{fig:planner_prompt_env_persona}
\end{figure*}

The prompt template of the planner for the \texttt{EnvPersona} dataset is shown in Fig.~\ref{fig:planner_prompt_env_persona} (yellow part). It includes the agent information (i.e., personality) and the environment description (i.e., specific situation), along with two example outputs to guide the output format. The green block shows one detailed example of the personality and the situation. For each data point, the personality and the situation should be different. 

\begin{figure*}
    \centering
    \includegraphics[width=0.99\textwidth]{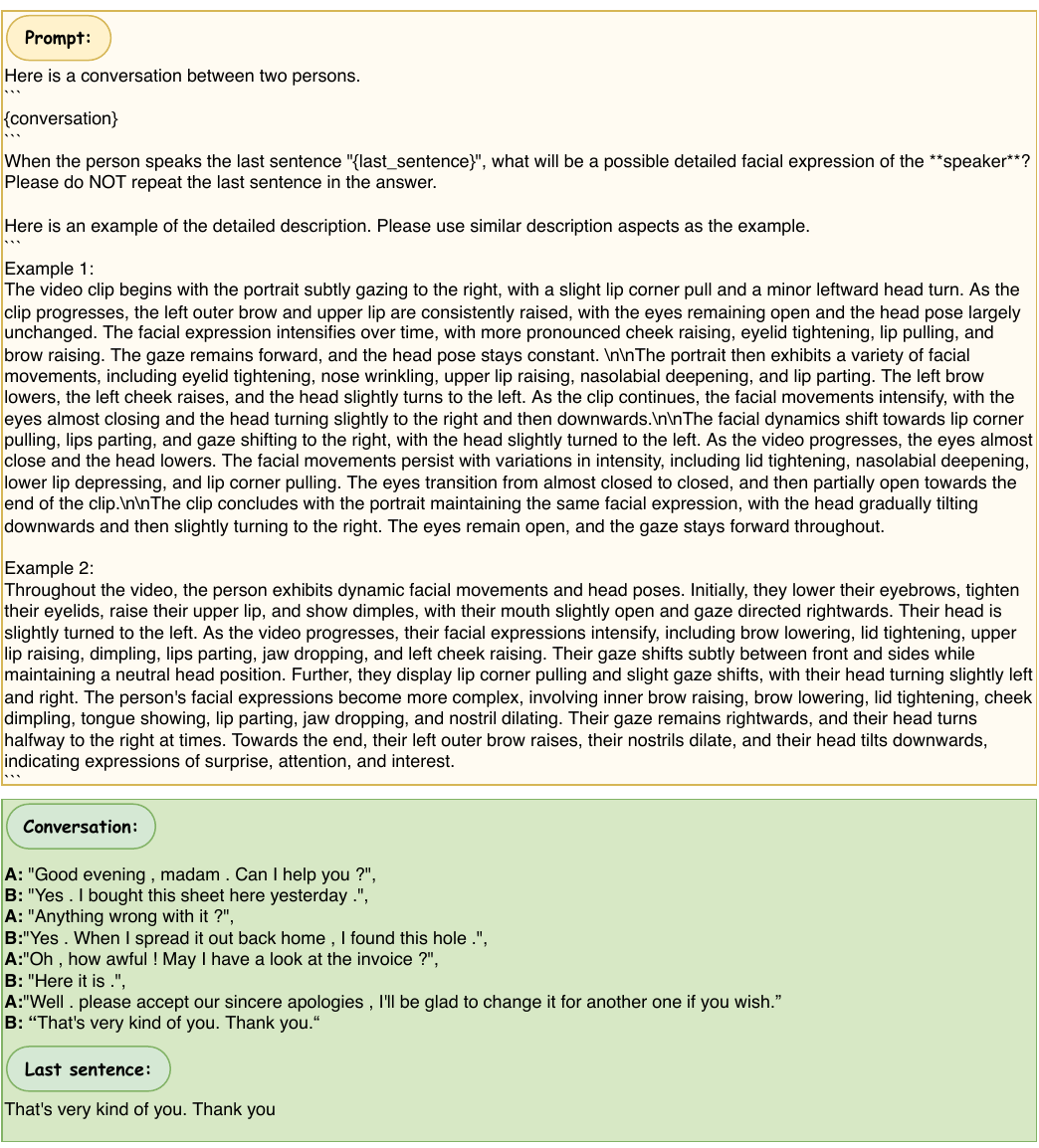}
    \caption{Planner prompt for the speaker in the \texttt{DailyDialogue} dataset.}
    \label{fig:planner_prompt_speaker}
    \vspace{20mm}
\end{figure*}

\begin{figure*}
    \centering 
    \includegraphics[width=0.99\textwidth]{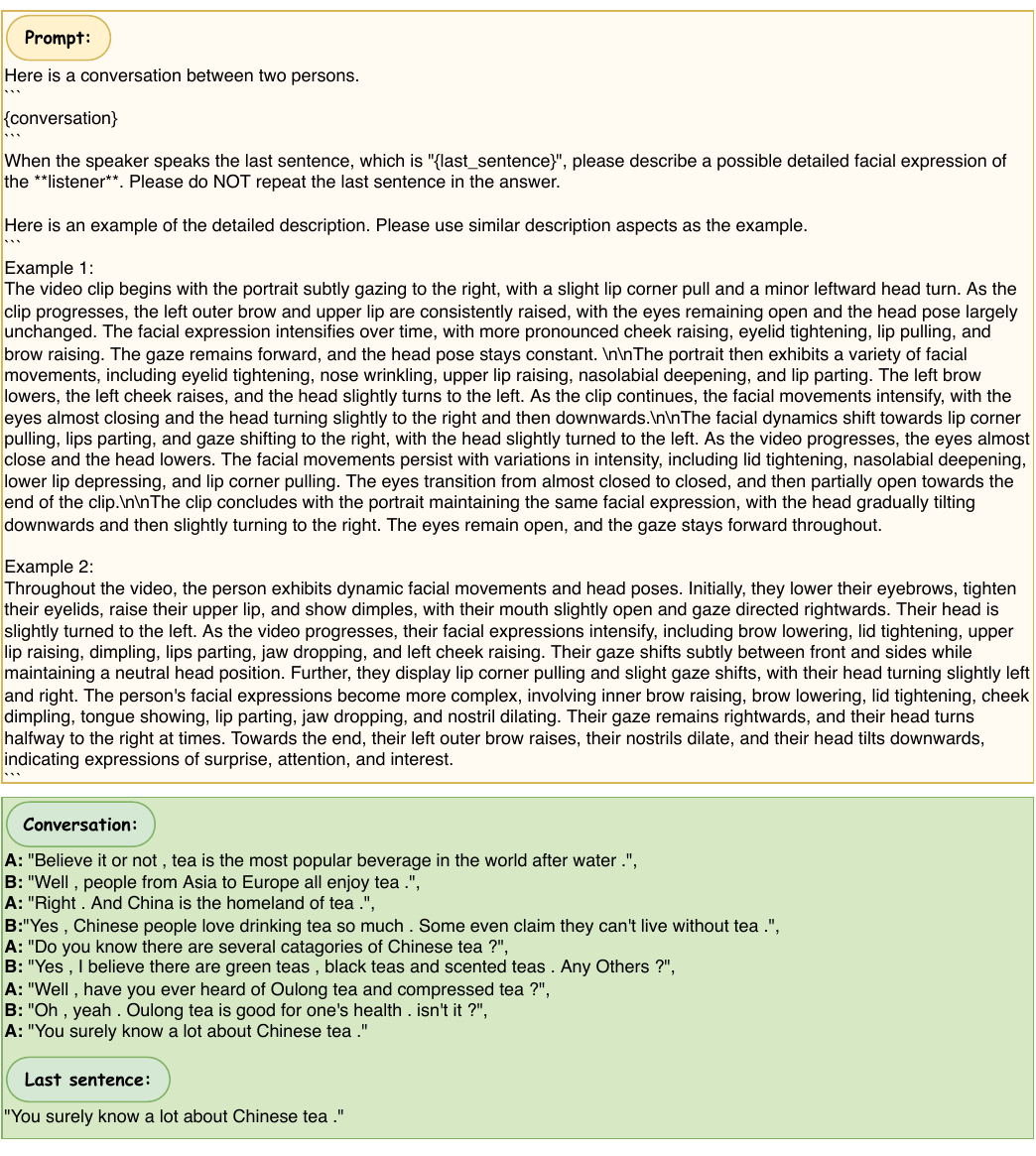}
    \caption{Planner prompt for the listener in the \texttt{DailyDialogue} dataset.}
    \label{fig:planner_prompt_listener}
\end{figure*}

The prompt template of the planner for the speaker and the listener of the \texttt{DailyDialogue} dataset are shown in Fig.~\ref{fig:planner_prompt_speaker} and \ref{fig:planner_prompt_listener} respectively. The template includes the conversation history and emphasizes the last sentence. Similarly, the green blocks demonstrate a detailed example.

\begin{table}\small 
    \centering
    \begin{tabular}{c|cc}
    \hline 
         & w. example & w.o. example \\
    \hline 
      EnvPersona-Agent   & \textbf{0.096} & 0.137 \\
      DailyDialogue-Listener & \textbf{0.129} & 0.148 \\
      DailyDialogue-Speaker &  \textbf{0.109} & 0.126 \\
    \hline 
    \end{tabular}
    \caption{FID between predicted descriptions and ground-truth descriptions.}
    \label{tab:fid}
\end{table}

The input for the planner is denoted as \texttt{<Inst, Env, Agent>}. We have demonstrated the utility of both \texttt{Env} and \texttt{Agent} qualitatively and quantitatively in facial expression planning. In this section, we underscore the importance of well-crafted instruction for the planner. To evaluate this, we experiment with excluding examples from the prompt during planning and examine the proximity of the generated expression descriptions to the ground truth, both with and without demonstration examples.

To be more specific, we extract text embeddings for each description using OpenAI's \texttt{text-embedding-ada-002} model. Subsequently, we compute the Fréchet Inception Distance (FID) score between the embeddings of the generated descriptions and the ground truth descriptions. Recognizing that text embeddings reside in a low-dimensional subspace where the covariance matrix can approach singularity, potentially causing issues in FID score calculation, we initially project the embeddings to a $20$-dimensional subspace using Principal Component Analysis (PCA). The FID score is then computed within the PCA space. The results are presented in Table~\ref{tab:fid}. Notably, the inclusion of demonstration examples consistently decreases the FID score across all three scenarios.

\subsection{EnvPersona Dataset Synthesis}
We first select $8$ emotions as seeds, including \emph{angry, disgust, contempt, fear, happy, sad, surprised, and neutral}, which we call the seed emotions. For each emotion, we ask ChatGPT to create $25$ diverse situations when a person may feel this emotion. So we can obtain a total of $25\times8=200$ different environment descriptions covering a person's daily life. Besides, we design $6$ persons with different personas. The main characters are \emph{brave, timid, calm, sensitive, pessimist and optimistic}. The detailed persona descriptions are enriched using ChatGPT given the main character. Thus we have a $200\times6=1200$ combination of (environment, persona) pairs.

\subsection{FilmData Generation Pipeline}
\begin{figure*}
	\centering
	\includegraphics[width=0.99\textwidth]{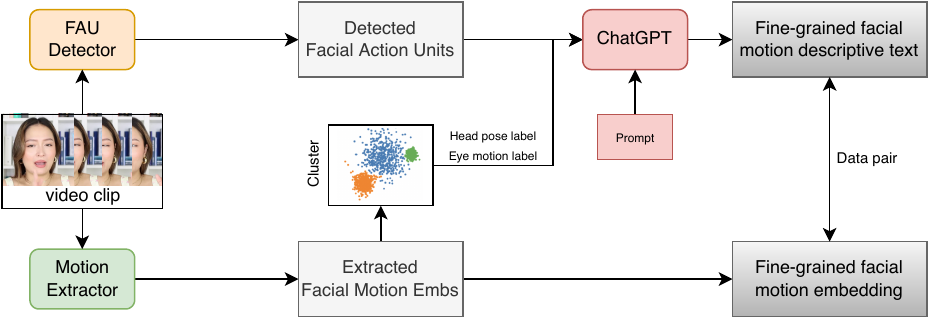}
	\caption{Our automatic data annotation pipeline.}
	\label{fig:data_pipeline}
\end{figure*}

\begin{figure*}
	\centering
	\includegraphics[width=0.99\textwidth]{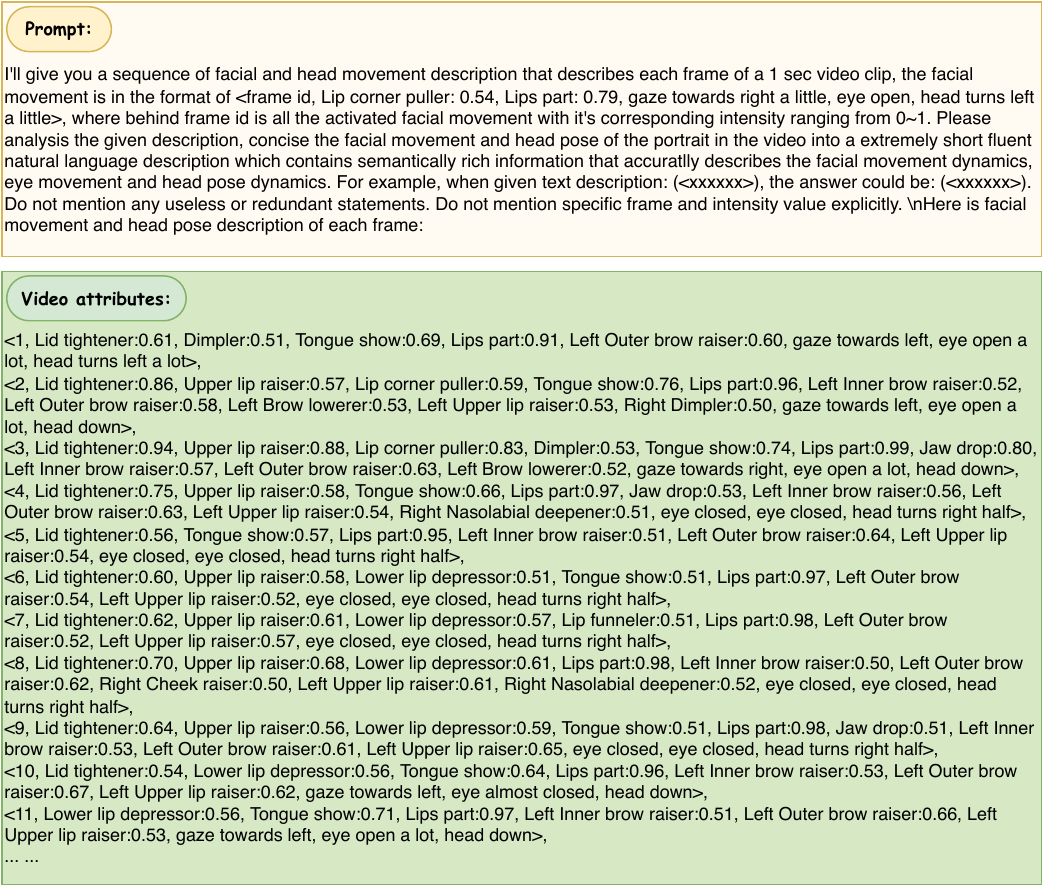}
	\caption{Pormpt for 1-second video description generation.}
	\label{fig:data_prompt}
\end{figure*}

\begin{figure*}
	\centering
	\includegraphics[width=0.99\textwidth]{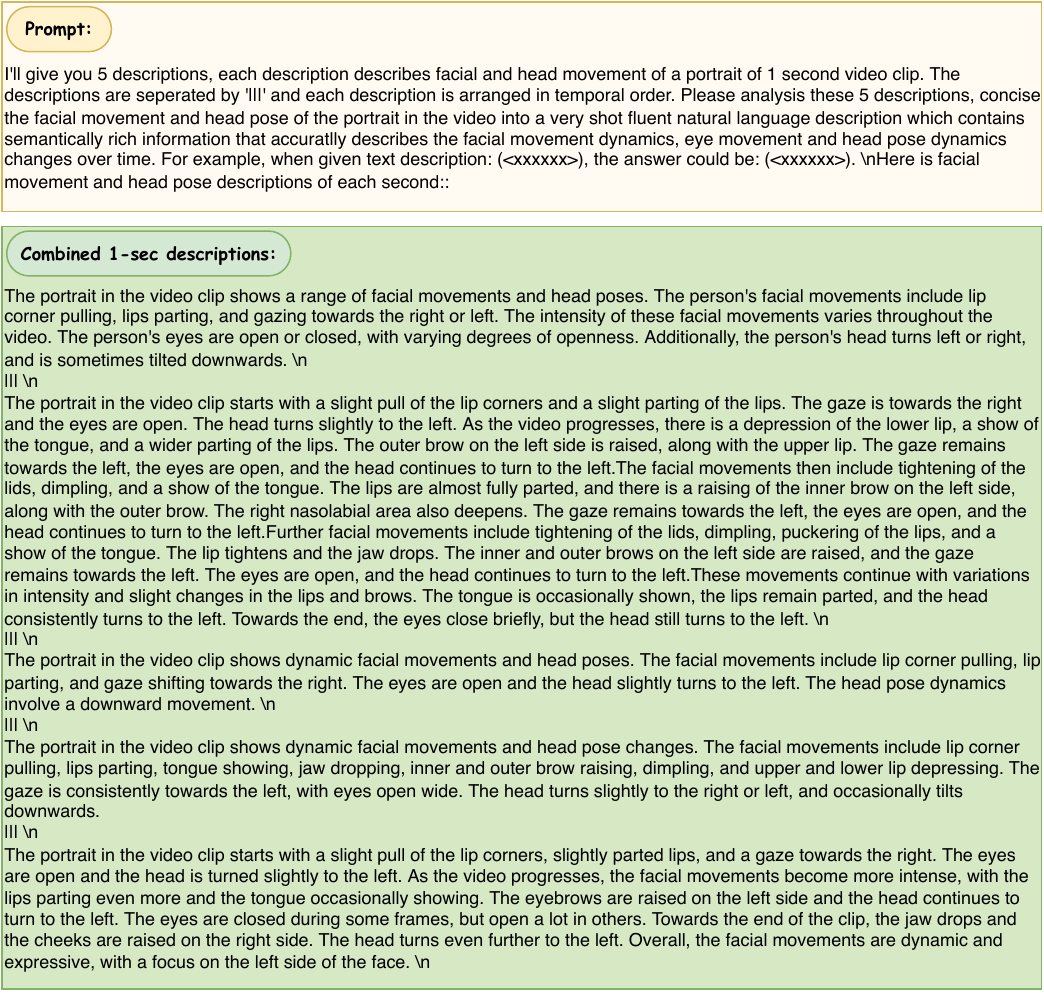}
	\caption{Pormpt for 5-second video description generation from 5 1-second videos.}
	\label{fig:data_prompt2}
\end{figure*}

In this section, we give a detailed implementation of how we build the \textbf{FilmData}, the overview of the data curation pipeline is illustrated in~\ref{fig:data_pipeline}. There are three main components in our data curation pipeline: \emph{Facial Action Unit Detection}, \emph{Motion Embedding Cluster}, and \emph{Attributes to Description Translation}.
\paragraph{Facial Action Unit Detection} We use~\cite{me_graphau} as the facial action unit detector, there is a pre-trained model trained on a hybrid dataset~\cite{zhang2014bp4d, mavadati2013disfa, Yan_2020_ACCV, kollias2018aff, yan2014casme, lucey2010extended}, results in a total of $41$ FAU labels. The FAU labels we used in FAU detection are shown in~\ref{tab:faus}. We only activate FAUs whose predicted probability is greater than $0.5$, and use the corresponding label to form \emph{video attributes} which will feed into ChatGPT together with a prompt to generate the final description.
\paragraph{Motion Embedding Cluster} The cluster labels we used for gaze, head pose, and eye blink are shown in~\ref{tab:cluster}. We take these clustered labels as complementary to FAUs. We only activate one class with maximal probability for each of the three attributes.
\paragraph{Attributes to Description Translation} We demonstrate the process of how the detected attributes are translated into fine-grained text motion description in Fig~\ref{fig:fine-description-generate} of the main paper, here we describe the detailed prompt design of 1-second clip description generation and the detailed prompt of longer clip description generation, shown in~\ref{fig:data_prompt} and~\ref{fig:data_prompt2}, respectively.

\begin{table*}[htbp]
    \centering
    \begin{tabular}{c|c|c|c|c|c}
    \toprule 
    \textbf{AU1} & Inner brow raiser     & \textbf{AU17} & Chin raiser             & \textbf{AUR1} & Right inner brow raiser \\
    \hline
    \textbf{AU2} & Outer brow raiser     & \textbf{AU18} & Lip pucker              & \textbf{AUL2} & Left outer brow raiser  \\
    \hline
    \textbf{AU4} & Brow lowerer          & \textbf{AU19} & Tongue show             & \textbf{AUR2} & Right outer brow raiser \\
    \hline
    \textbf{AU5} & Upper lid raiser      & \textbf{AU20} & Lip stretcher           & \textbf{AUL4} & Left brow lowerer       \\
    \hline
    \textbf{AU6} & Cheek raiser          & \textbf{AU22} & Lip funnelerr           & \textbf{AUR4} & Right brow lowerer      \\
    \hline
    \textbf{AU7} & Lid tightener         & \textbf{AU23} & Lip tightener           & \textbf{AUL6} & Left cheek raiser       \\
    \hline
    \textbf{AU9} & Nose wrinkler         & \textbf{AU24} & Lip pressor             & \textbf{AUR6} & Right cheek raiser       \\
    \hline
    \textbf{AU10} & Upper lip raiser     & \textbf{AU25} & Lips part               & \textbf{AUL10} & Left upper lip raiser    \\
    \hline
    \textbf{AU11} & Nasolabial deepener  & \textbf{AU26} & Jaw drop                & \textbf{AUR10} & Right upper lip raiser    \\
    \hline
    \textbf{AU12} & Lip corner puller    & \textbf{AU27} & Mouth stretch           & \textbf{AUL12} & Left nasolabial deepener  \\
    \hline
    \textbf{AU13} & Sharp lip puller     & \textbf{AU32} & Lip bite                & \textbf{AUR12} & Right nasolabial deepener \\
    \hline
    \textbf{AUL4} & Left brow lowerer    & \textbf{AU38} & Nostril dilator         & \textbf{AUL14} & Left dimpler              \\
    \hline
    \textbf{AU15} & Lip Corner depressor & \textbf{AU39} & Nostril compressor      & \textbf{AUR14} & Right dimpler   \\
    \hline
    \textbf{AU16} & Lower lip depressor  & \textbf{AUL1} & Left inner brow raiser  & - & - \\
    \bottomrule 
    \end{tabular}
    \vspace*{-2mm}
    \caption{Lables of facial action units.}
    \label{tab:faus}
   \vspace*{-4mm}
\end{table*}

\begin{table*}[htbp]
    \centering
    \begin{tabular}{c|c|c|c|c|c}
    \toprule 
    \multicolumn{2}{c}{Gaze} & \multicolumn{2}{c}{Head pose} & \multicolumn{2}{c}{Eye blink} \\
    \hline
    \textbf{G1} & Gaze towards up a little    & \textbf{H1} & Head turns left a lot     & \textbf{E1} & Eye open a lot    \\
    \textbf{G2} & Gaze towards left a little  & \textbf{H2} & No head pose              & \textbf{E2} & Eye almost closed \\
    \textbf{G3} & Gaze towards ahead          & \textbf{H3} & Head turns left half      & \textbf{E3} & Eye closed        \\
    \textbf{G4} & Gaze towards down a little  & \textbf{H4} & Head turns right half     & \textbf{E4} & Eye open a lot    \\
    \textbf{G5} & Gaze towards right          & \textbf{H5} & Head up a little          & \textbf{E5} & Eye open          \\
    \textbf{G6} & Eye closed                  & \textbf{H6} & Head down                 & - & - \\
    \textbf{G7} & Gaze towards left           & \textbf{H7} & Head turns left a little  & - & - \\
    \textbf{G8} & Gaze towards right a little & \textbf{H8} & Head turns right a lot    & - & - \\
    \textbf{G9} & Eye open                    & \textbf{H9} & Head turns right a little & - & - \\
    \bottomrule 
    \end{tabular}
    \vspace*{-2mm}
    \caption{Additional cluster labels.}
    \label{tab:cluster}
   \vspace*{-4mm}
\end{table*}

\subsection{Motion Tokenization}
Our motion embedding including expression embedding and head pose embedding extracted from images using~\cite{Wang_2023_CVPR}, and is then tokenized to a sequence of expression tokens and a sequence of head pose tokens. The tokenization enables the alignment of motion space with language space, and then the motion token can be predicted by a Large Language model. We use two VQ-VAEs to discretize expression embedding and head pose embedding independently. We train the two VQ-VAEs with the same model architecture and hyperparameters. Take expression VQ-VAE as an example, it includes an encoder network, a decoder network, and a codebook that consists of $512$ embeddings $\textbf{C}\in \mathbb{R}^{512\times d_{c}}$, where $d_{c}$ is the dimension of the embeddings. Each embedding in the codebook represents a discretized token after tokenization. The input expression embedding sequence $E=(\textbf{e}_{1}, \textbf{e}_{2}, ..., \textbf{e}_{t})$ is first normalized by mean and std computed across the training set, and are then fed into the encoder, with the latent embedding $Z=(\textbf{z}_{1}, \textbf{z}_{2}, ..., \textbf{z}_{t})$ as output, where $\textbf{z}_{i}\in\mathbb{R}^{d_{c}}$. Then $Z$ is tokenized by the \emph{quantizer} to one of the embeddings in the codebook as follows:
\begin{equation}
\label{eq:quantizer}
\small
\begin{aligned}
Q(z_{i}) = \arg\min_{1<j<N}{\lVert z_{i} - c_{j}\rVert^2},
\end{aligned}
\end{equation}
Where $N$ represents the size of the codebook, which is $512$ here. The tokenized embedding $C=(\textbf{c}_{1}, \textbf{c}_{2}, ..., \textbf{c}_{t})$ can then be decoded by the decoder and re-normalized by mean and std to generate the reconstructed expression embeddings $\hat{E}=(\Hat{\textbf{e}}_{1}, \Hat{\textbf{e}}_{2}, ..., \Hat{\textbf{e}}_{t})$.

\section{More Results}\label{sec:results}
\subsection{Qualitative Results of Planner}

\begin{figure*}
    \centering
    \includegraphics[width=0.96\textwidth]{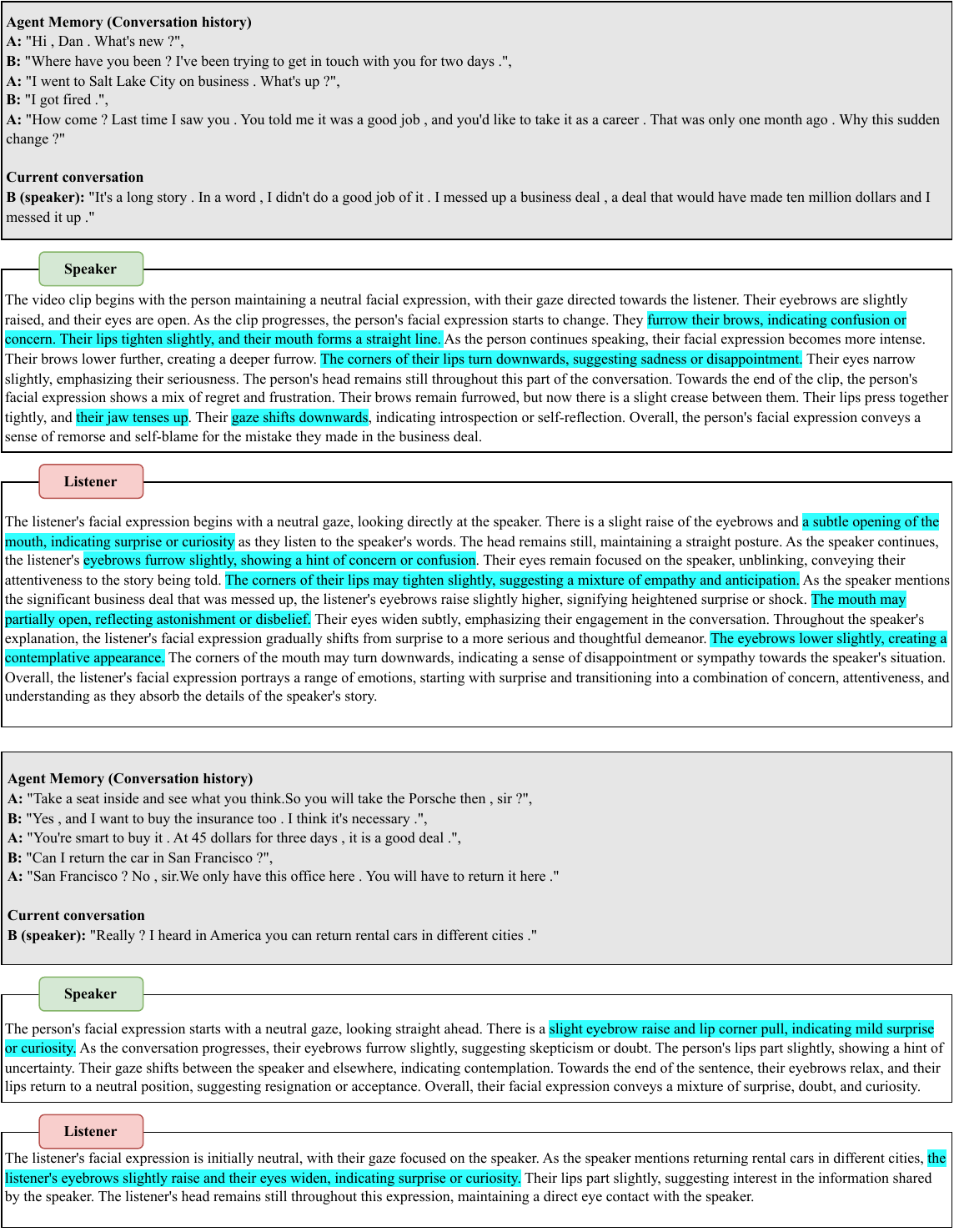}
    \caption{Qualitative results of planner on the \texttt{DailyDialogue} dataset.}
    \label{fig:planner_gallary_dailydialogue}
\end{figure*}

\begin{figure*}
    \centering
    \includegraphics[width=0.98\textwidth]{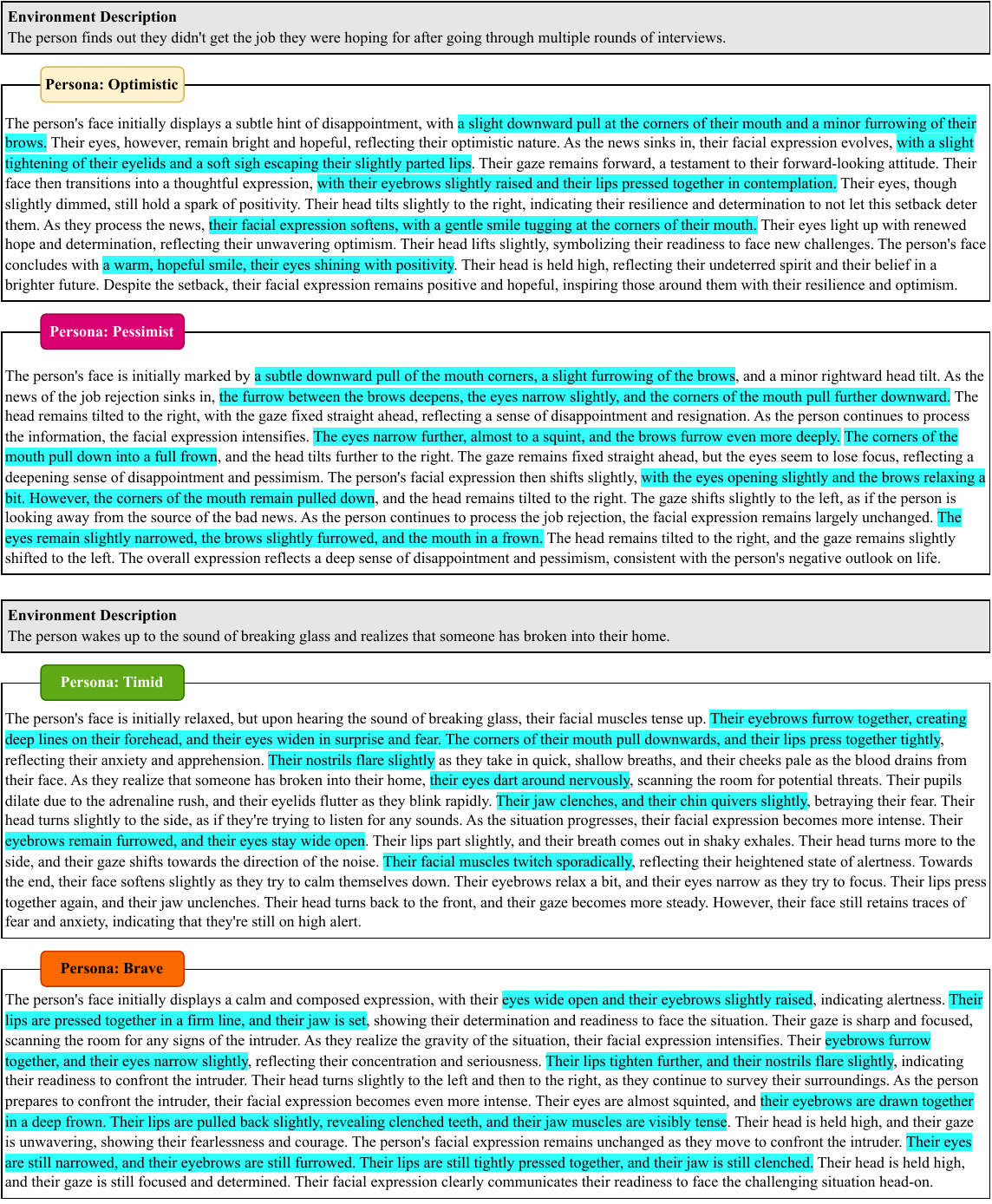}
    \caption{Qualitative results of planner on the \texttt{EnvPersona} dataset.}
    \label{fig:planner_gallary_env}
\end{figure*}

We show more qualitative results similar to Fig.~\ref{fig:planner_gallery} in the main paper and we provide full generated descriptions in this section. Two examples on the \texttt{DailyDialogue} dataset are demonstrated in Fig.~\ref{fig:planner_gallary_dailydialogue}. Both the facial expressions of the speaker and the listener are provided. The key expressions are highlighted in blue background. We also illustrate two examples on the \texttt{EnvPersona} dataset in Fig.~\ref{fig:planner_gallary_env}. For each example, we show two persons with different personas. In the first example, a calm person tends to have mild expressions like \emph{no noticeable tightening of the eyelids, a mild frown}. For a brave person, the expression becomes more intense, such as \emph{a more pronounced furrowing of the brows, a firm set of the jaw, and a steady gaze}. In the second case, where the person wakes up to the sound of breaking glass and realizes that someone has broken into their home,  a timid person's expression shows nervousness, fear, and anxiety while an optimistic person's expression shows less worry.

\subsection{Qualitative Results of Driving Engine}
\begin{figure*}
	\centering
	\includegraphics[width=0.99\textwidth]{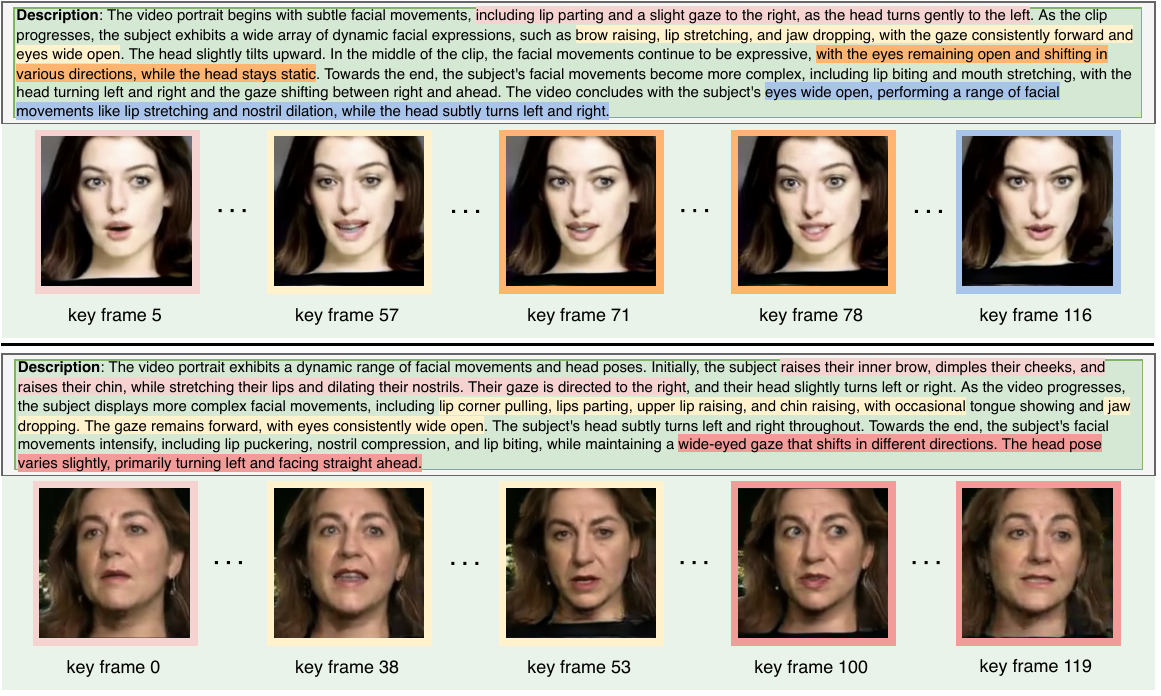}
	\caption{Qualitative results of driving engine with different identity rendered, note that the generated videos include lip motion from speech audio, so there may be variant lip shapes.}
	\label{fig:driving_gallery1}
\end{figure*}

\begin{figure*}
	\centering
	\includegraphics[width=0.99\textwidth]{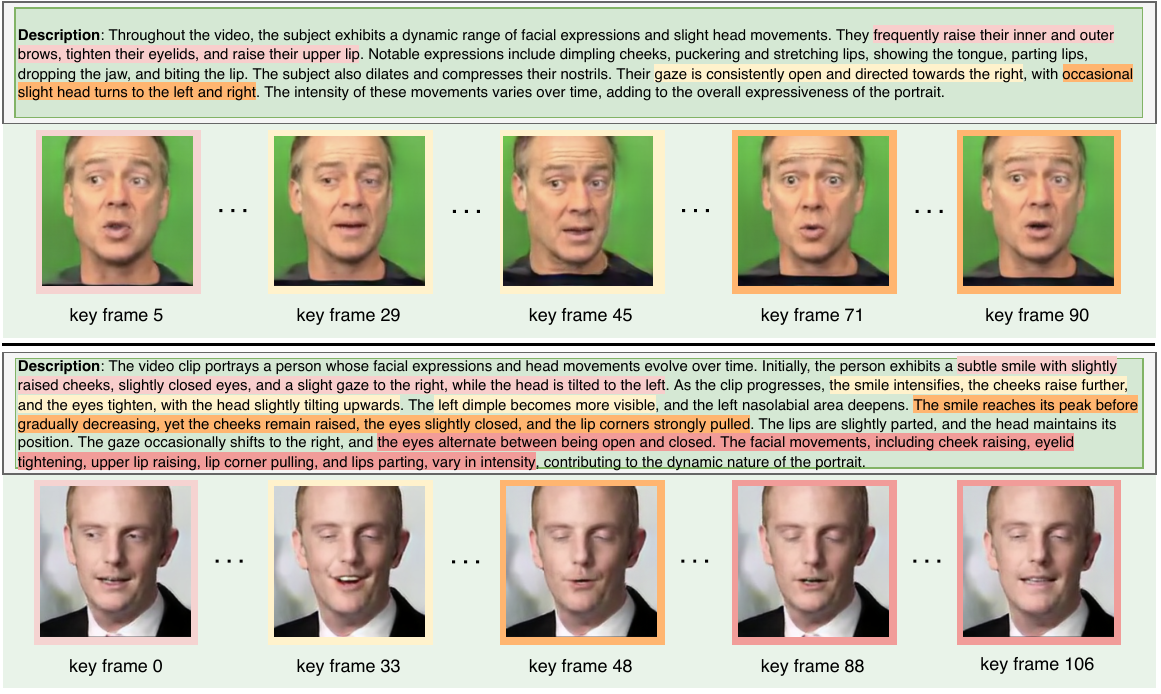}
	\caption{Qualitative results of driving engine with different identity rendered, note that the generated videos include lip motion from speech audio, so there may be variant lip shapes.}
	\label{fig:driving_gallery2}
\end{figure*}

We present more visual results of different identities driving for driving engine in Fig.~\ref{fig:driving_gallery1} and~\ref{fig:driving_gallery2}. We also provide a video demo, please refer to the video supplementary.


\subsection{Analysis for System Results}
Tab.~\ref{tab:user_study} in the main paper reports \emph{Hit@3} for both datasets, where the number of the randomly selected additional samples is $9$. We report more results including \emph{Hit@1} and \emph{Hit@2} in Tab.~\ref{tab:system_results}. Compared with Tab.~\ref{tab:planner_hit_at_k} in the main paper, the hit rate is quite low, we argue that this is caused by a long end-to-end evaluation pipeline, which brings accumulated error, especially the low rendering quality and the existence of description gap. The low rendering quality will decrease the FAUs detection accuracy, and thus affect the accuracy of generated motion descriptions. The description gap between planner generated one and training data also contributes to the accumulated error. On the one hand, this decreases the accuracy of the driving engine, on the other hand, this will also bring extra unfavorable factors to the ranking, such as a certain degree of mismatch by using different description words. 

However, the imperfect metric does not actually mean the low accuracy of our system. Please refer to the demo video to see the actual effect.

\subsection{Compared with Other Methods}
We compare our method with the latest yet most relevant work ~\cite{Learn2Listen_LLM_2023_ICCV} and conduct a user study to evaluate from three aspects: Naturalness, Variance, and Matchness. Here we describe the details of the user study.

We invited eight people with no prior knowledge about agents or avatars as the subject of our investigation. We show ten randomly chosen comparison videos, each video shows the result of ours and the result of~\cite{Learn2Listen_LLM_2023_ICCV}, together with the corresponding speech audio. We also show the corresponding context of the conversation and the transcript of the speech content. We ask each user to evaluate these two methods from the three aspects we mentioned above and score from $0$ to $4$, in which $0$ is the lowest score and $4$ is the highest score. For example, if one thinks method \emph{A} shows completely naturalness, then he/she should score \emph{A} a $4$. Then we compute the winning rate for each aspect, the winning rate is computed as follows:

\begin{equation}
\label{eq:win_rate}
\small
\begin{aligned}
W_{m1} = \frac{1}{N+M}\sum_{1<i<N}{\sum_{1<j<M}{\mathbbm{1}\{S_{m1}^{ij}>S_{m2}^{ij}}\}},
\end{aligned}
\end{equation}
where $N$ is the number of subjects, $M$ is the number of videos, $S_{m}^{ij}$ is score of $j-th$ video of method $m$ from $i-th$ subject. We neglect the item when the score is equal.
\paragraph{Analysis for Results} Our system outperforms~\cite{Learn2Listen_LLM_2023_ICCV} on \emph{Naturalness} by 15\%, indicating that the motion pattern generated by our system is more similar to the ground truth. We also outperform on \emph{Variance} and \emph{Matchness}, because our system planner generates motion sequences based on different environments and personalities, which is more compatible with different contexts. On the contrary, ~\cite{Learn2Listen_LLM_2023_ICCV} is a person-specific method, one pre-trained model can only mimic the motion pattern of one specific person, which is limited for a generalized system.



\section{Limitation and Future work}\label{sec:limitation}
Even though our systems show promising results, there are several noticeable limitations that need to be addressed in future works.

\subsection{Time Alignment Between Text and Motion}
Like many other text-to-video works, the text descriptions are not strictly aligned with real motion in the generated video. Although the descriptions are in time order, they only describe the facial movements during a specific time period. We contend that our system could achieve the alignment across different snippets specified by ``\emph{At the beginning of the video..}", ``\emph{As the video progresses...}" and ``\emph{Towards the end of the video...}",  it, however, can not guarantee the local time alignment within one specific time snippet. In fact, it could be very difficult to control motion in a specific timestamp through a text description, due to the fact that a very long description will exceed the maximal token limitation if we describe each frame in detail. We plan to improve this in future work.
\subsection{Time Alignment Between Speech and Motion}
Although we don't explicitly model the speech speed into the current planner, it's technically feasible to make it part of the agent persona and let the planner predict the speed based on the dynamic context. Preparing a person-specific training dataset would be an interesting direction to push this effort.
\subsection{Rendering Quality}
In this paper, our primary focus is to verify the overall disentangled design principle, meanwhile showcasing the potential of leveraging LLM for detailed facial motion planning, which is an important leap step toward a fully autonomous avatar agent.  For the sake of clarity and simplicity, we utilized a pre-trained disentangled talking face model as the render, namely PD-FGC~\cite{Wang_2023_CVPR}. Using more advanced 3D-aware high-resolution rendering and integrating it into our proposed pipeline would be an interesting future work.

\end{document}